\documentclass[ journal]{IEEEtran} 
\usepackage{mwe}

\usepackage{graphicx}
\usepackage{graphbox}
\usepackage{caption}
\usepackage{subcaption}
\usepackage[utf8]{inputenc}
\usepackage{amsmath}
\usepackage{mathtools}
\usepackage{float}
\usepackage{svg}
\usepackage{soul}
\usepackage{amsthm}
\usepackage{cite}
\usepackage{xcolor}
\usepackage{url}
\usepackage{comment}
\newtheorem{theorem}{Theorem}
\newtheorem{proposition}{Proposition}
\newtheorem{corollary}{Corollary}
\newtheorem{definition}{Definition}
\newtheorem{lemma}{Lemma}
\newtheorem*{remark}{Remark}
\newcommand{\etal}{\emph{et al}.\@ }
\usepackage{multicol}
\usepackage{dblfloatfix} 
\usepackage{here}
 \usepackage[normalem]{ulem} 
\IEEEoverridecommandlockouts

\begin{document}

\title{Passive Realizations of Series Elastic Actuation: Effects of Plant and Controller Dynamics on\\ Haptic Rendering Performance}

% You will get a Paper-ID when submitting a pdf file to the conference system
\author{Celal~Umut~Kenanoglu   \hspace{60mm}  Volkan~Patoglu\thanks{C.~U.~Kenanoglu is with the Munich Institute of Robotics and Machine Intelligence at Technical University of Munich, 80992, Germany. V.~Patoglu is with the Faculty of Engineering and Natural Sciences at Sabanci University, Istanbul, 34956, Turkiye. C.~U.~Kenanoglu’s work was carried out during his graduate studies at Sabancı University.\\% \hspace{40mm} 
E-mail:~\{umut.kenanoglu,volkan.patoglu\}@sabanciuniv.edu}}

%\author{\authorblockN{Michael Shell}
%\authorblockA{School of Electrical and\\Computer Engineering\\
%Georgia Institute of Technology\\
%Atlanta, Georgia 30332--0250\\
%Email: mshell@ece.gatech.edu}
%\and
%\authorblockA{Twentieth Century Fox\\
%\and
%\authorblockN{James Kirk\\ and Montgomery Scott}
%\authorblockA{Starfleet Academy\\
%San Francisco, California 96678-2391\\

% avoiding spaces at the end of the author lines is not a problem with
% conference papers because we don't use \thanks or \IEEEmembership

%Homer Simpson\authorrefmark{2},
%Montgomery Scott\authorrefmark{3} and
%Eldon Tyrell\authorrefmark{4}}
%Telephone: (800) 555--1212, Fax: (888) 555--1212}
%\authorblockA{\authorrefmark{4}Tyrell Inc., 123 Replicant Street, Los Angeles, California 90210--4321}}

\maketitle

\begin{abstract}
We introduce minimal passive physical equivalents of series (damped) elastic actuation (S(D)EA) under closed-loop control to determine the effect of different plant parameters and controller gains on the closed-loop performance of the system and to help establish an intuitive understanding of the passivity bounds. Furthermore, we explicitly derive the feasibility conditions for these passive physical equivalents and compare them to the necessary and sufficient conditions for the passivity of S(D)EA under velocity sourced impedance control (VSIC) to establish their relationship. Through the passive physical equivalents, we rigorously compare the effect of different plant dynamics (e.g., SEA and SDEA)  on the system performance. We demonstrate that passive physical equivalents make the effect of controller gains explicit and establish a natural means for effective impedance analysis. We also show that passive physical equivalents promote co-design thinking by enforcing simultaneous and unbiased consideration of (possibly negative) controller gains and plant parameters. We demonstrate the usefulness of negative controller gains when coupled to properly designed plant dynamics.  Finally, we provide experimental validations of our theoretical results and characterizations of the haptic rendering performance of S(D)EA under VSIC. 

\end{abstract}
\vspace{-1mm}


\IEEEpeerreviewmaketitle

\vspace{-3mm}
\section{Introduction}

%\PARstart{E}{stablishing} safe and natural physical human-robot interactions (pHRI) is vital in many applications, including collaborative manufacturing, service, surgical, assistive, and rehabilitation robotics. 

\PARstart{S}{afe} and natural physical human-robot interactions (pHRI)  necessitate precise control of the impedance characteristics of the robot at the interaction port~\cite{colgate_hogan_1988}. Series elastic actuation (SEA) is a commonly employed interaction control paradigm that has been introduced in~\cite{howard1990joint,pratt1995series,robinson1999series} to address the fundamental trade-off between the stability robustness and the control performance of closed-loop force control systems~\cite{eppinger_seering,newman_1992,kenanoglu2023}. SEA relies on an intentionally introduced compliant element between the actuator and interaction ports and utilizes the model of this compliant element to implement closed-loop force control. Thanks to SEA, the strict stability bounds on the controller gains induced due to sensor-actuator non-collocation and actuator bandwidth limitations can be relaxed, leading to high stability robustness and good rendering performance.  On the negative side, the compliant element significantly decreases the system bandwidth; consequently, the control effort increases quickly for high-frequency interactions, resulting in actuator (velocity and/or torque) saturation. %an_hollerbach, hogan_1985, 

Series damped elastic actuation (SDEA) extends SEA by introducing a linear viscous dissipation element parallel to the series elastic element~\cite{oblak2011design,hurst2004series,garcia2011combining,laffranchi2011compact,kim2017enhancing}. SDEA not only helps increase the force control bandwidth of SEA~\cite{hurst2004series} but also provides additional advantages, in terms of improving energy efficiency~\cite{garcia2011combining}, reducing undesired oscillations~\cite{laffranchi2011compact}, alleviating the need for derivative control terms~\cite{kim2017enhancing}, and relaxing the upper-bound on passively renderable stiffness~\cite{kenanoglu2023}.

Given S(D)EA is generally designed as lumped-parameter LTI systems, the coupled stability of interactions with S(D)EA has commonly been studied through passivity analyses~\cite{colgate_hogan_1988}. %Since inanimate environments are passive and non-malicious human interactions do not intentionally aim to destabilize a system, the passivity analysis can be utilized to conclude coupled stability of pHRI~\cite{colgate_hogan_1988}. 
 While passivity conditions are known to be conservative,  closed-form analytical passivity conditions derived through such analysis are informative,  as they provide insights into how system parameters affect stability robustness.

Stability robustness and rendering performance have been established to conflict with each other; therefore, there exist trade-offs involved in the design of series (damped) elastic actuation (S(D)EA). A clear understanding of these trade-offs is crucial for safe and high-fidelity renderings. In this study,  we demonstrate that passive physical (mechanical) equivalents are instrumental in understanding the effect of different plant parameters and controller gains on closed-loop haptic rendering performance and passivity bounds. Furthermore, we show passive physical equivalents enable symbolic comparisons of the performance of different plant dynamics (e.g.,  SEA vs SDEA) on passive haptic rendering performance; if continuity is established among realizations, the effect of each controller term on closed loop system dynamics of different plants can be rigorously studied.

%\vspace{.7mm}
%\smallskip 
\emph{Contributions:} 
We propose (i)~the use of \emph{passive physical equivalents} as an informative means of providing physical insight into the passivity-performance trade-offs of S(D)EA, and (ii)~derive minimal passive physical equivalents of S(D)EA under closed-loop control, with their feasibility region. We advocate for passive physical equivalents as they
\vspace{-.3mm}
\begin{itemize}%[leftmargin=*]
\item  provide sufficient conditions for the passivity bounds,
\item establish an intuitive understanding of the passivity bounds by explicitly highlighting the contribution of each plant parameter and controller gain on the rendering performance,
\item make the authority of controller terms on the plant dynamics explicit,
%\item \textcolor{blue}{motivate the use of negative controller gains},
\item  do not distinguish between the plant and controller parameters,  promoting co-design of S(D)EA by enforcing simultaneous and unbiased consideration of these (possibly negative) parameters to improve system performance, %\textcolor{blue}{as demonstrated for Voigt model rendering with SEA under VSIC},
\item subsume the effective impedance analysis that decomposes the output impedance into its basic mechanical primitives and extend it by providing an explicit topological connection of these fundamental elements, and
\item enable rigorous comparisons of the effect of different plant/controller dynamics (e.g., SEA and SDEA)  on the haptic rendering performance. %\sout{and controller architectures  (e.g., P-P and P-PI)}
\end{itemize}

Furthermore, (iii)~we rigorously study the passivity of S(D)EA to establish closed-form analytical solutions for the \emph{necessary and sufficient conditions} for the passivity of S(D)EA while rendering Voigt models, springs, and the null impedance. 

Our results significantly extend the previously established bounds in the literature, as we provide passivity bounds for Voigt model rendering with S(D)EA under VSIC, prove the necessity and sufficiency of these conditions, and allow for negative controller gains. Our study not only establishes the feasibility of passive Voigt model rendering with SEA under VSIC through the use of negative controller gains but also demonstrates that such a control approach can be practicable through a co-design, where physical damping is intentionally added to the plant dynamics, for instance via electrical damping. It also provides the necessary and sufficient conditions for the passivity of SDEA while rendering Voigt models. 

Overall, the derivation of minimal passive physical equivalents of S(D)EA that are close to the open-loop plant dynamics and that lend themselves to simple interpretations, ensuring continuity among various realizations of different plants, determination of the control authority on plant parameters, and insightful discussions of closed-loop performance via passive physical equivalents are among our novel contributions. 

\vspace{-3mm}
\section{Related Work}

%In this section, we discuss the related works on frequency domain passivity of S(D)EA, and review the classical results and recent developments in the realization of passive networks.

%There are robotic applications which rely on open-loop force/impedance control which eliminate negative impacts of force sensor. End-effector force/impedance is directly translated to the motor torques/impedance for open-loop systems. However, transparency of mechanical design is critical to provide sufficient performance for open-loop system. Alternatively, closed-loop force control is used in many robotic applications to compensate parasitic force from the mechanical design. However, it is a challenging problem to achieve coupled stability for closed-loop force control because of unavoidable bounds on the controller gains due to non-collocation and bandwidth limitations of sensors and actuators which enforce a fundamental trade-off between stability robustness and control performance~\cite{an_hollerbach,eppinger_seering,newman_1992}.
\vspace{-1mm}
\subsection{Passivity Analysis of SEA} \vspace{-1mm}

Pratt~\etal have presented the first passivity analysis for SEA, and provided sufficient conditions for a SEA under a filtered PID force controller with a feedforward compensator~\cite{pratt1995series}. Since the introduction of velocity-sourced impedance control (VSIC) for SEA~\cite{howard1990joint,pratt2004,Wyeth2006,Wyeth2008}, the passivity of SEA under VSIC has been  studied extensively~\cite{vallery2007passive,vallery2008compliant,tagliamonte2014passivity,calanca2017impedance,Tosun2020,kenanoglu2023,Kenanoglu2023b}. 
VSIC has become the most popular force controller for SEA, as its cascaded control architecture with an inner motion control loop can effectively eliminate parasitic forces ---undesired effects due to dissipation, compliance, and inertial dynamics that negatively affect the rendering transparency, leading to a linear system and  good rendering performance~\cite{howard1990joint,robinson1999series,Wyeth2008,Otaran2021}. Furthermore, VSIC is easy-to-use, since this controller does not rely on the dynamic model of the plant and the controller gains can be tuned empirically. 

Vallery~\etal have provided a set of sufficient conditions for null impedance and linear spring rendering with SEA under VSIC~\cite{vallery2007passive,vallery2008compliant}.  They have also proved that the passively renderable stiffness of a SEA under VSIC is upper bounded by the physical stiffness of the compliant element of SEA~\cite{vallery2008compliant}.  Tagliamonte~\etal have provided less conservative sufficient conditions for the passivity of SEA under VSIC during null impedance, linear stiffness, and Maxwell body rendering~\cite{tagliamonte2014passivity}. They have also proved that Voigt model \emph{cannot} be passively rendered with SEA under VSIC when the controllers are PI-PI and the controller gains are positive. % and shown that the maximum renderable stiffness not only depends on the physical stiffness, but also the physical damping in the system. 
 Calanca~\etal have presented sufficient conditions for the passivity of SEA under four different control architectures: VSIC, basic impedance, collocated admittance, and collocated impedance controllers~\cite{calanca2017impedance}. They have shown that the passively renderable virtual stiffness of all of these  control architectures is also limited by the physical stiffness of the compliant element~\cite{calanca2017impedance}. Calanca~\etal have also advocated for the use of acceleration feedback to compensate for the load dynamics~\cite{calanca2017impedance,Calanca2018_ff}. While acceleration feedback can help improve performance~\cite{pratt1995series,robinson1999series,Calanca2018_ff}, the fundamental passive stiffness rendering limitations of SEA cannot be relaxed, as long as the controllers are kept causal~\cite{kenanoglu2023}.
 
 %and an impedance controller with perfect acceleration feedback has been recommended. Theoretically, ideal acceleration feedback can be used to cancel out the influence of load dynamics; however, noise and bandwidth restrictions of acceleration signals, as well as the potential for overestimation of feed-forward signals resulting in feedback inversion, are significant practical challenges that have restricted the adaptation of acceleration-based control, since initially proposed in~\cite{pratt1995series,robinson1999series}.   

Tosun and Patoglu~\cite{Tosun2020} have presented the necessary and sufficient conditions for the passivity of SEA under VSIC for null impedance and linear spring rendering. The earlier sufficiency bounds on controller gains have been relaxed and the range of impedances that can be passively rendered has been extended in this study. Furthermore, it has been shown that the integral gain of the motion controller is required to render stiffness if the force controller utilizes an integral term. 

Authors have proposed model reference force control (MRFC) for SEA and provided a passivity analysis of this control scheme, under model mismatch. In particular, sufficient conditions for the passivity of SEA under MRFC during null impedance rendering have been presented in~\cite{Kenanoglu2022}. 

Recently, authors have established a fundamental limitation of passive spring rendering with SEA, by proving that the physical stiffness of its compliant element cannot be exceeded with any (linear or nonlinear) causal controller~\cite{kenanoglu2023}. Authors have also studied the effect of low-pass filtering on the passivity and rendering performance SEA under VSIC~\cite{Kenanoglu2023b}.

This study significantly extends the passivity results for SEA under VSIC in~\cite{vallery2007passive,vallery2008compliant,tagliamonte2014passivity,calanca2017impedance,Tosun2020,kenanoglu2023,Kenanoglu2023b} by deriving the necessary and sufficient conditions for the passivity of SEA under VSIC during Voigt model rendering. It establishes that passive rendering of Voigt models with SEA under VSIC is feasible when negative controller gains are utilized and demonstrates the practical application of such renderings via addition of damping to the plant. This study also presents novel minimal passive physical equivalents of SEA under VSIC during the Voigt model, linear spring, and null impedance rendering.

\vspace{-3mm}
\subsection{Passivity Analysis of SDEA}

SDEA generalizes SEA by introducing a viscous dissipation element parallel to the series elastic element.
Accordingly, the passivity analysis of SDEA also generalizes the passivity analysis of SEA. However, passivity analysis of SDEA has received relatively less attention in the literature, since the resulting closed-form solutions of these systems are more complex and much harder to interpret~\cite{focchi2016robot,oblak2011design,Mengilli2020,mengilli2021passivity}.

The passive range of virtual stiffness and damping parameters for SDEA under a cascaded impedance controller with an inner torque loop acting on a velocity-compensated plant and load dynamics have been studied in~\cite{focchi2016robot}. In this controller, a positive velocity feedback loop provides velocity compensation by attempting to extend the bandwidth of the torque control loop under passivity constraints. 

Oblak and Matjacic~\cite{oblak2011design} have conducted a passivity analysis of SDEA under an unconventional basic impedance controller. In this controller, a force sensor is employed after the end-effector inertia to measure the interaction forces, and these forces are used for closed-loop force control, in addition to the series damped elastic element. It has been demonstrated that a sufficient level of mechanical damping is required in the compliant element  to ensure the passivity of linear stiffness rendering using this control architecture. Furthermore, sufficient conditions to passively render linear springs have been proposed, which include a lower bound on the required level of physical damping.

Mengilli~\etal have presented sufficient conditions for the passivity of SDEA under VSIC for the null impedance, linear spring, and Voigt model rendering~\cite{mengilli2021passivity}. They have demonstrated that thanks to the damping of the compliant element, passive spring renderings with SDEA can exceed the physical stiffness of the compliant element. They have extended their results to absolute stability and two-port passivity analyses and derived the necessary and sufficient conditions for appropriate virtual couplers~\cite{Mengilli2020}. In~\cite{kenanoglu2023}, authors have studied  the necessary and sufficient conditions for the  passivity of linear spring rendering with SDEA under a cascaded controller that neglects the forces induced on the damping element.

This study significantly extends the one-port passivity results in~\cite{Tosun2020} by extending them to SDEA, and the results in~\cite{mengilli2021passivity,kenanoglu2023} by establishing the necessity bounds for SDEA under VSIC and generalizing the results to include negative controller gains. This study also presents novel minimal passive physical equivalents of SDEA under VSIC  to provide intuition about the passivity bounds and to study the trade-offs involved in the rendering performance.

%The biggest problem of SEA is decreased large force bandwidth caused by the increase of the sensor compliance under actuator saturation \cite{pratt1995series}. Therefore, series damped elastic actuation (SDEA) has been studied in several works~\cite{hurst2004series,oblak2011design,garcia2011combining,laffranchi2011compact,kim2017enhancing} which can increase the force control bandwidth \cite{hurst2004series}. SDEA has additional advantages over SEA such as energy efficiency \cite{garcia2011combining}, reduction of undesired oscillations \cite{laffranchi2011compact}, and lack of need for D-control terms \cite{kim2017enhancing}, they do not study the coupled stability of interaction with SDEA during impedance rendering. 

\vspace{-3mm}
\subsection{Realization of Passive Physical Equivalents}

Passive physical equivalents are studied in the field of network synthesis, which aims to rigorously describe physically realizable behaviors in a given domain with specified components. %In particular, the goal of network synthesis is to design a passive network of fundamental elements to realize a given driving-point impedance.
 Colgate and Hogan have advocated the use of passive physical equivalents for the analysis of contact instability observed in interaction control~\cite{Colgate_realizations}. They have studied uncontrollable elements under all causal controllers and through passive mechanical realizations of force-controlled systems, they have demonstrated a fundamental limitation on inertia compensation under passivity constraints for force-feedback systems with sensor-actuator non-collocation. They have also illustrated that the passive physical equivalents promote the use of negative controller gains and a simultaneous consideration of the design of mechanical and controller subsystems.

%\textcolor{blue}{Ortega~\etal have presented  physical equivalents of linear systems to motivate  passivity-based control through intuitive and easy-to-follow examples~\cite{ortega2013passivity}.}

  \begin{figure*}[t!]
  \centering
  \includegraphics[width=0.825\linewidth]{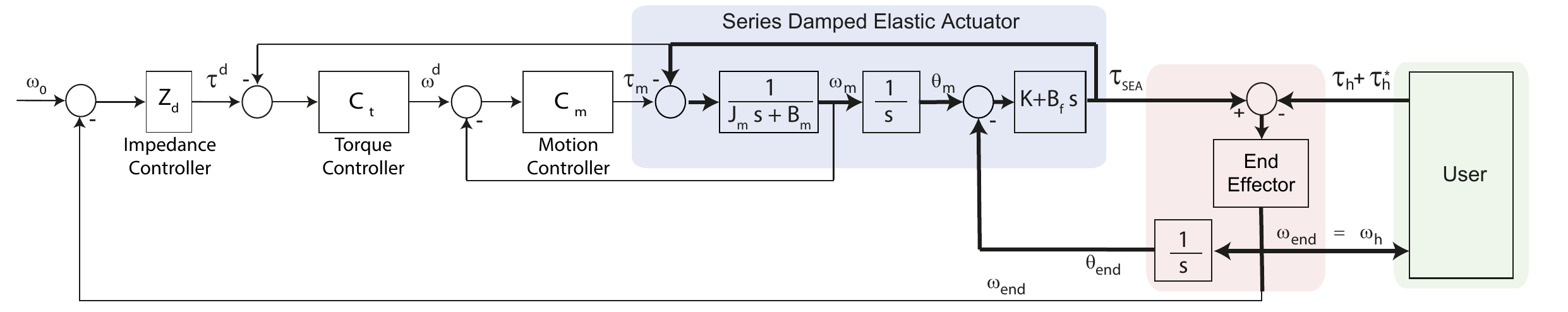}
    \vspace{-0.65\baselineskip}
    \caption{Block diagram of S(D)EA under VSIC}
    \label{fig:VSICclosed}      \vspace{-1.3\baselineskip}
\end{figure*}

Inspired by~\cite{Colgate_realizations}, this paper focuses on passive mechanical realizations of S(D)EA under VSIC. Extending the methods in~\cite{Colgate_realizations}, our linear passive mechanical networks are built utilizing springs, dampers, and \emph{inerters}---a relatively recently introduced fundamental element of the mechanical domain~\cite{smith2002synthesis,Chen_missing}. The use of the inerter element is crucial as it completes the force-current analogy between the electrical and mechanical domains by introducing an ideal linear two-terminal energy storage element equivalent to an ungrounded capacitor. The completion of the analogy has a major impact, as it enables all of the previously established results in the electrical network synthesis to be equivalently expressed in the mechanical domain. Thanks to this analogy, all classical results, including Foster's reactance theorem~\cite{foster1924reactance} characterizing lossless networks, Brune's construction method~\cite{brune1931synthesis} for the minimal realization of general positive-real functions using resistors, inductors, capacitors, and transformers, and Bott-Duffin theorem~\cite{bott1949impedance} indicating transformers are not necessary for the synthesis of positive-real impedances, can be directly used for the network synthesis in the mechanical domain.

While network synthesis in the electrical domain has received much attention during the era of analog circuits, the diminished attention has been renewed during the last decade, especially in the mechanical domain, with the introduction of inerter element and demonstration of its successful applications in the design of passive  suspensions~\cite{smith2002synthesis,chen2009restricted,chen2012realization,chen2013realization,chen2013realizations,chen2015realization}. %,chen2007mechanical

Kalman has also advocated for a renewed focus on network synthesis to establish a general theory of the subject, pointing out the high potential of this field to have a wide impact in a broad range of applications~\cite{Kalman2010}. Accordingly, recent results have been established to extend the classical ones.
Chen and Smith have studied the most general class of mechanical impedances that can be realized using one damper, one inerter, and an arbitrary number of spring elements while allowing no levers~\cite{chen2009restricted}. Jiang and Smith have studied the realizability conditions for positive-real biquadratic impedance functions which can be realized by 
%five-element~\cite{jiang2011realization} and 
 six-element~\cite{jiang2012realization} networks. Chen~\etal have extended their earlier results in~\cite{chen2013realization} and established the realizability conditions to two special class of mechanical networks: networks with biquadratic functions with an extra pole at the origin~\cite{chen2013realizations} and networks that are constituted of one inerter, one damper, and at most three springs~\cite{chen2015realization}. 
Hughes and Smith have extended the classical results on Bott–Duffin realization procedure by discussing the minimality and uniqueness of these realizations among all series-parallel networks realizing biquadratic minimum functions~\cite{Hughes_2014_realization}. Hughes has further extended these results and established minimal network realizations for the class of impedances realized by series–parallel networks containing at most three energy storage elements~\cite{Hughes_2020}. %Hughes_2017,
 Morelli and Smith have presented an enumerative approach to the passive network synthesis and provided a classification for networks of restricted complexity~\cite{Morelli_book_2019}. Readers are referred to the survey by Hughes~\etal for a review of the recent developments~\cite{Hughes_survey_2018}. 

To the best of the authors' knowledge, this is the first study in which passive mechanical equivalents are systematically used to analyze passivity-performance trade-offs of S(D)EA. Furthermore, extending the methods utilized by the seminal work in~\cite{Colgate_realizations}, this study introduces inerter elements to the analysis of interaction control systems.

\vspace{-3mm}
\subsection{Rendering Performance} \label{sec:PerformanceAnalysisofSDEA}

While the coupled stability of pHRI systems constitutes an imperative design criterion, the rendering performance of the system is also significant for natural interactions. Transparency is a commonly used concept in the evaluation of the haptic rendering performance, as it quantifies the match between the mechanical impedance of the virtual environment and the impedance felt by the user, with the requirement of identical force/velocity responses~\cite{Hannaford1989, Hashtrudi-Zaad2001}. $Z_{\text{width}}$ is another commonly used metric that quantifies the difference between the minimum and the maximum passively renderable impedances~\cite{Colgatezwidth94}. 

Given that the rendered impedance is a function of frequency, both of these metrics are also quantified as such; however, the frequency dependence of these metrics make their interpretation challenging. To provide physical intuition to the characteristics of the impedance at the interaction port, it is common practice to decompose the impedance into its basic \emph{mechanical primitives} through effective impedance analysis~\cite{mehling2005increasing,colonnese2015rendered,tokatli2015stability,tokatli_patoglu_2018,aydin2018stable,Yusuf_multi_2020}. In particular, the effective impedance definitions partition the frequency-dependent impedance transfer function into its real and imaginary parts and assign the real positive part to effective damping, while the imaginary part is mapped to effective spring and effective inertia components based on the phase response of the impedance. %Along these lines, the impedance is decomposed into its \emph{mechanical primitives}. 

This study proposes passive mechanical equivalents to significantly extend the effective impedance analysis since a feasible realization also provides a topological connection of the fundamental mechanical elements. It demonstrates that passive physical equivalents subsume the effective impedance analysis and provide a more intuitive understanding of system behavior through its underlying components.

\vspace{-2mm}
\section{Preliminaries} \vspace{-1.5mm}

%In this section,  we present the linear time-invariant (LTI) model of SDEA under VSIC and review the relevant theorems used to study the passivity in the frequency domain.

%  \smallskip
%\vspace{-3mm}
\subsection{System Description}

Consider a single-axis SDEA plant without its controller. Let the reflected inertia of the actuator be denoted by $J_m$, the viscous friction of the actuator including the reflected motor damping is denoted by $B_m$, and the physical compliant element and viscous damper, arranged in parallel between the end-effector and the actuator, be denoted by $K$ and $B_f$, respectively. Let $\omega_m$ and $\omega_{end}$ denote the actuator and end-effector velocities and $\tau_m$ be the actuator torque. 
  
%\begin{figure}[h!]
%  \centering
%  \includegraphics[width=.7\linewidth]{figures/SchematicSDEA.pdf}
%  \vspace{-.3\baselineskip}
 %   \caption{Schematic representation of SDEA (The plant reduces to an SEA when $B_f=0$.)}
  %  \vspace{-3mm}
  %  \label{fig:SEA_SDEA_sch}
  %\end{figure}

The torque $\tau_{sea}$ on the damped compliant element, also called the physical filter, is equal to the sum of the torques induced on the linear spring and the viscous damper elements. The plant reduces to a SEA, when $B_f$ is set to zero. In this case, $\tau_{sea}$ can be computed using the deflections of the linear spring~$K$, according to the Hooke's law. 

The human interaction is modeled with two components:  $\tau_h$ represents the passive component of the applied torques while $\tau_h^*$ is the deliberately applied active component that is assumed to be independent of the system states~\cite{colgate_hogan_1988}. We assume that the non-malicious human interactions do not intentionally aim to destabilize the system. It is considered that the end-effector inertia of SDEA is negligible or is a part of the user dynamics such that $\tau_{sea}(s) \approx \tau_h + \tau_h^*$; hence, the impedance at the interaction port is defined as $Z_{out}(s) = -\frac{\tau{sea}(s)}{\omega_{end}(s)}$, where the spring-damper torque is considered positive in compression.  
  
Figure~\ref{fig:VSICclosed} depicts the block diagram of SDEA under VSIC, where the thick lines represent physical forces. In VSIC, the inner velocity control loop of the cascaded controller renders the system into an ideal motion source and acts on references generated by the outer torque control loop to keep the spring-damper deflection at the desired level to match the reference force. The symbols  $C_t$ and $C_m$ denote the torque and motion (velocity) controllers, respectively.

%\smallskip
%\vspace{.25mm}

The following assumptions are considered for the analysis: %\vspace{-.5mm}
\begin{itemize}
    \item A lumped-parameter LTI model is considered; nonlinear effects, such as backlash and saturation are neglected.
    \item Electrical dynamics are neglected and actuator velocity is assumed to be available with a negligible time delay.
    \item The deflection of the physical filter and its time derivative are assumed to be measured with a  negligible~delay.
    \item Without loss of generality, a zero motion reference  ($\omega_d$$=$$0$) is assumed for the virtual environment and the transmission ratio is set to one for simplicity.
    \item The physical plant parameters are assumed to be positive, while the controller gains can be negative, as long as the inner motion control loop is asymptotically stable.
\end{itemize}

\vspace{-4mm}
\subsection{Passivity Theorems}

The passivity of an LTI network is equivalent to the positive realness of its impedance transfer function $Z(s)$~\cite{colgate_hogan_1988}. The positive realness of a rational function $Z(s)$ with real coefficients can be studied according to Theorem~\ref{theorem:positivereallness} as follows.  \vspace{-2mm}
\begin{theorem}[\!\!\cite{colgate_hogan_1988,Haykin70}] %,kuorealness
A rational LTI impedance transfer function $Z(s)$ with real coefficients is passive if and only if:\\
(1)~Z(s) has no poles in the right half plane, and\\
(2)~$Re[Z(jw)] \ge 0$ for  $w \in (-\infty,\infty)$, and\\
(3)~Any poles of Z(s) on the imaginary axis are simple with positive and real residues.  \label{theorem:positivereallness}
\end{theorem} \vspace{-2mm}

%\vspace{-1mm}
The following useful lemmas have been established in the literature to determine the necessary and sufficient conditions for
the passivity of LTI systems. \vspace{-1mm}

\begin{lemma}
Let $Z(s)=N(s)/D(s)$ be an impedance transfer function. Then, $Re[Z(jw)] \ge 0$ iff the test polynomial $P(w) \ge 0$ for any value of $w$, where $P(w)=Re[N(jw) D(-jw)] = \sum_{i=0}^{n} d_i w^i$, and $d_i$ represents the coefficient of $w^i$.
\label{lemma:positivereal}
\end{lemma}\vspace{-2.5mm}

\begin{lemma}
Let $f(s)=a_3 s^3 + a_2 s^2 + a_1 s + a_0$ for $a_i \ge 0$ be the third-order characteristic equation of a  system. Then, f(s) has no roots in the open right half plane if and only if $a_3 \ge 0$, $a_2 \ge 0$, $a_0 \ge 0$, and $a_1 a_2 - a_0 a_3 \ge 0$.  If these inequalities are strictly greater than zero, then the system has no roots on the imaginary axis.
\label{lemma:stability}
\end{lemma}\vspace{-2.5mm}

\begin{lemma}[\!\!\cite{Mengilli2020}]
A polynomial of the form $p(x)=p_2 x^2 + p_1 x + p_0$, $p(x) \ge 0$ for all $x \ge 0$ if and only if $p_2 \ge 0$, $p_0 \ge 0$  and $p_1\geq-2\sqrt{p_0p_2}$.
\label{lemma:sturm}
\end{lemma}

\vspace{-4mm}
\subsection{Passive Physical Equivalents and Inerter}

%\vspace{-5mm}

\begin{definition}
Passive physical equivalents describe physically realizable behaviors with a passive network of fundamental elements in a  domain to realize a driving-point impedance. 
\end{definition} \vspace{-1mm}

In the force-current analogy between the mechanical and electrical domains, forces are considered to be analogous to currents, while velocities are analogous to voltages. Passive mechanical networks are built utilizing springs, dampers, and inerters. The inerter is an ideal energy storage element that completes the force-current analogy between the mechanical and electrical domains~\cite{smith2002synthesis,Chen_missing}. The interter element generalizes the more familiar mass/inertia element in the mechanical domain, which is analogous to the restricted case of a grounded capacitor in the electrical domain. %Particularly, an inerter is an ideal linear two-terminal energy storage element in the mechanical domain that is equivalent to an ungrounded capacitor in the electrical domain. 

\begin{definition}
An inerter is an ideal linear two-terminal energy storage element in the mechanical domain with terminal forces proportional to the relative acceleration between them. %
\end{definition}  \vspace{-3mm}

%Figure~\ref{fig:forcecurrent} presents the force-current analogy between fundamental two-terminal elements in both domains.

%\begin{figure}[h]
 % \centering
 % \includegraphics[width=.725\linewidth]{figures/MecvsElecv2.pdf}
 % \vspace{-.55\baselineskip}
 %   \caption{Force-current analogy between the fundamental %two-terminal elements in the electrical and mechanical %domains.}
 % \vspace{-1.25\baselineskip}
 %   \label{fig:forcecurrent}
 % \end{figure}

\begin{table*}[t!]
  \centering
    \caption{Passive physical equivalents for SDEA and SEA under VSIC} 
      \vspace{-.75\baselineskip} 
  \includegraphics[keepaspectratio=true, width=.78\linewidth]{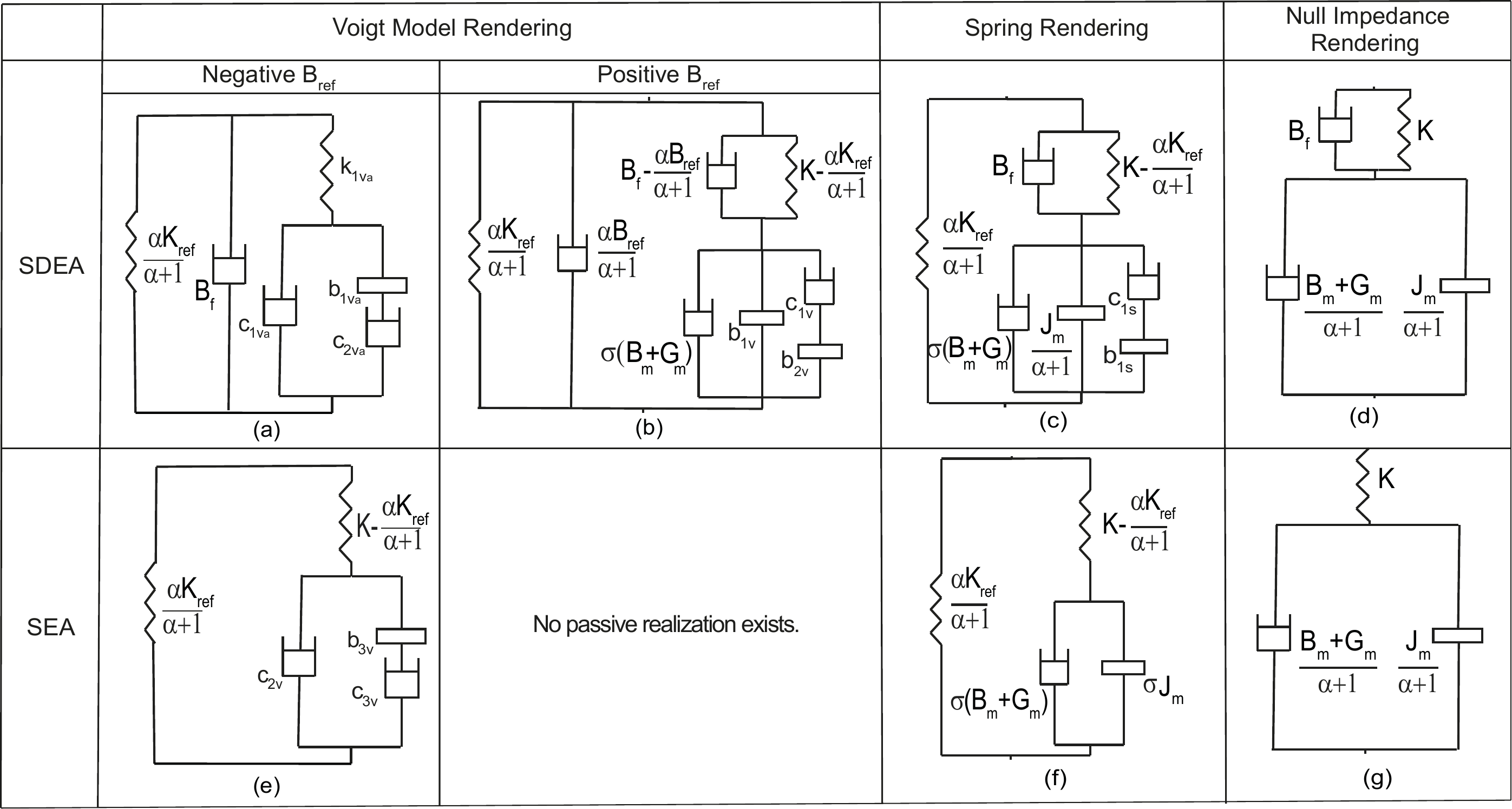}
    \label{fig:allresults}
    \vspace{-2.1\baselineskip}
\end{table*}

%%%% VOIGT MODEL RENDERING 
%%%%%%%%%%%%%%%%%%%%%%%%%%%%%%%%%%%%%%%%%%%%%%%%%%%%%%%%%%%%%%%%%%%%%%%%%%%%%%%%%%%%%%%%%%%%%%%%%%%%%%%%%%%%%%%%%%%%%%%%%%%%
%%%%%%%%%%%%%%%%%%%%%%%%%%%%%%%%%%%%%%%%%%%%%%%%%%%%%%%%%%%%%%%%%%%%%%%%%%%%%%%%%%%%%%%%%%%%%%%%%%%%%%%%%%%%%%%%%%%%%%%%%%%%
%%%%%%%%%%%%%%%%%%%%%%%%%%%%%%%%%%%%%%%%%%%%%%%%%%%%%%%%%%%%%%%%%%%%%%%%%%%%%%%%%%%%%%%%%%%%%%%%%%%%%%%%%%%%%%%%%%%%%%%%%%%%
%%%%%%%%%%%%%%%%%%%%%%%%%%%%%%%%%%%%%%%%%%%%%%%%%%%%%%%%%%%%%%%%%%%%%%%%%%%%%%%%%%%%%%%%%%%%%%%%%%%%%%%%%%%%%%%%%%%%%%%%%%%%
%%%%%%%%%%%%%%%%%%%%%%%%%%%%%%%%%%%%%%%%%%%%%%%%%%%%%%%%%%%%%%%%%%%%%%%%%%%%%%%%%%%%%%%%%%%%%%%%%%%%%%%%%%%%%%%%%%%%%%%%%%%%
%%%%%%%%%%%%%%%%%%%%%%%%%%%%%%%%%%%%%%%%%%%%%%%%%%%%%%%%%%%%%%%%%%%%%%%%%%%%%%%%%%%%%%%%%%%%%%%%%%%%%%%%%%%%%%%%%%%%%%%%%%%%
 
%\vspace{-3mm}
\section{Passive Physical Equivalents of S(D)EA}
\label{sec:SEApassivity}

In this section, we present the passive physical equivalents of SDEA and SEA under VSIC with proportional~(P) controllers while rendering Voigt and linear spring models,  with their feasibility analysis. We study haptic rendering performance during rendering Voigt and spring models through passive physical equivalents. We show the relationship between effective impedance and passive physical equivalents.

\vspace{-3mm}
\subsection{Voigt Model Rendering}

In this subsection, we present passive physical equivalents of SDEA/SEA during Voigt model rendering, with their corresponding feasibility conditions.

\medskip
\subsubsection{Series Damped Elastic Actuation~(SDEA)}
When both the torque and velocity controllers of VSIC are proportional, the impedance at the interaction port of SDEA under VSIC during Voigt model rendering is $Z_{Voigt}^{SDEA^{P{\text -}P}}(s) =$ \vspace{-3mm}

\begin{equation} \small
\!\!\frac{ \splitfrac {B_f \,J_m \,s^3 +{\left[B_f (\,B_m +\,G_m) +J_m \,K+B_{ref} \,B_f \, \alpha \right]}\,s^2} {+{\left[ K \, (B_m +G_m +B_{ref} \, \alpha )+B_f  \,K_{ref} \, \alpha \right]}\,s+ K\,K_{ref}\,\alpha} }{J_m \,s^3 +{\left[B_m +G_m +B_f \, (1+\alpha) \right]}\,s^2 + K \, {\left(1+\alpha \right)}\,s}
    \label{eqn:SDEAvoigtPP}
\end{equation} \normalsize \vspace{-3mm}

\noindent where $\alpha=G_m G_t$.  

\smallskip
\noindent \underline{Passive Physical Equivalent of SDEA under VSIC:} A minimal realization of Eqn.~(\ref{eqn:SDEAvoigtPP}) characterizing SDEA under VSIC during Voigt model rendering when both controllers are P and $B_{ref} > 0$ is presented in Table~\ref{fig:allresults}b, where the parameters of this physical realization can be listed as \vspace{-4mm}

\footnotesize
\begin{eqnarray} %\vspace{-3mm}
b_{1v} \! &=& \!\frac{J_m}{(\alpha+1)^2} \, - \frac{\alpha}{(\alpha+1)^2} \, \frac{\left(B_{ref} - B_f \right) }{ B_f  }\, J_m \nonumber \\
c_{1v} \! &=& \!\frac{\alpha}{(\alpha+1)^2} \, \frac{{\left(B_f \,K_{ref} \!-\! B_{ref} \,K \right)}\,{\left[B_f \, (B_m + G_m) \!-\!J_m \,K\right]}}{{B_f }^2 \,K } \nonumber \\ 
b_{2v} \!&=& \!\frac{\alpha}{(\alpha+1)^2} \, \frac{{\left(B_f \,K_{ref} \!-\! B_{ref} \,K \right)}\,{\left[B_f \, (B_m + G_m) \!-\!J_m \,K\right]}}{B_f \,K^2 }  \nonumber
\end{eqnarray} \normalsize  \vspace{-4mm}

\noindent with $\sigma=\frac{1}{\alpha+1} - \frac{\alpha}{(\alpha+1)^2} \frac{K_{ref}}{K}$. \vspace{1mm}

For the realization in Table~\ref{fig:allresults}b to be feasible, all components of the model should be non-negative. First of all, $\frac{\alpha}{\alpha+1 }B_{ref}$ and $\frac{\alpha}{\alpha+1 } K_{ref}$ should be positive to have a feasible realization in Table~\ref{fig:allresults}b while rendering Voigt models. When $G_m$ and $G_t$ are non-negative, the non-negativeness of these terms imposes \vspace{-\baselineskip}

\small
\begin{eqnarray}  \vspace{-\baselineskip}
    K &\ge& \frac{\alpha  }{\alpha +1} \, K_{ref} \label{eqn:SDEAspringPP_spring} \\
    B_f &\ge& \frac{\alpha  }{\alpha +1} \, B_{ref} \label{eqn:SDEAspringPP_damping} \\
   \frac{B_f}{K} &\ge& \frac{J_m}{(B_m+G_m)}    \label{eqn:SDEAspringPP_extraterm}  
\end{eqnarray}  \normalsize 

Substituting Eqn.~(\ref{eqn:SDEAspringPP_damping}) into $(B_f \, K_{ref} - B_{ref} \, K)$ and invoking Eqn.~(\ref{eqn:SDEAspringPP_spring}), one can prove that $(B_f \, K_{ref} - B_{ref} \, K) \ge 0$. Eqns.~(\ref{eqn:SDEAspringPP_spring})~and~(\ref{eqn:SDEAspringPP_damping}) impose the upper bounds on passively renderable stiffness and damping levels. 

In Eqn.~(\ref{eqn:SDEAspringPP_extraterm}), $\frac{J_m}{(B_m+G_m)}$  captures the time constant of the motion-controlled mass-damper system, while $\frac{B_f}{K}$ is the time constant of the serial physical filter. Accordingly, the condition in Eqn.~(\ref{eqn:SDEAspringPP_extraterm}) imposes the intuitive constraint that the motion-controlled mass-damper model of the plant should respond faster than the interaction forces filtered by the physical filter, such that the system can keep up with these inputs to adequately cancel out the undesired dynamics and superpose the virtual impedance to be rendered.

Table~\ref{fig:allresults}a presents a minimal realization of Eqn.~(\ref{eqn:SDEAvoigtPP}) characterizing SDEA under VSIC during Voigt model rendering when both controllers are $P$ and $B_{ref} < 0$. This realization is presented as it, not only complements the realization in Table~\ref{fig:allresults}b for negative $B_{ref}$ values but also ensures continuity with the realization of SEA under VSIC during Voigt model rendering. The elements of this realization are prohibitively more complicated and only analyzed in the next section for the relatively simpler case of SEA.

\medskip
\noindent \underline{Haptic Rendering Performance through Realization:} 
The physical realization of SDEA under VSIC during Voigt model rendering in Table~\ref{fig:allresults}b  indicates three main branches in parallel: a spring-damper pair $\frac{\alpha}{(\alpha+1)}\,K_{ref}$--$\frac{\alpha}{(\alpha+1)}\,B_{ref}$ in parallel that converges to the Voigt model to be rendered, and a branch capturing the parasitic dynamics which are governed by a complex structure of damper-inertance terms that is connected to the system through a coupling filter that operates in series. %Note that there exist upper bounds on passively renderable stiffness and damping levels.

The coupling filter consists of a spring-damper pair in parallel, where the stiffness and damping of the filter are given by $K-\frac{\alpha}{(\alpha+1)} \, K_{ref}$ and $B_f-\frac{\alpha}{(\alpha+1)} \, B_{ref}$, respectively. The coupling filter indicates that the parasitic dynamics become more decoupled from the system as the control gains $G_t$ and $G_m$  increase. Furthermore, given that the coupling filter terms need to be positive, upper bounds are imposed on $K_{ref}$ and $B_{ref}$ of the Voigt models that can be passively rendered. 

The parasitic dissipation effects are split into two parts: a damper term scaled by $\sigma = \frac{1}{\alpha+1} - \frac{\alpha}{(\alpha+1)^2} \frac{K_{ref}}{K}$ indicating a significant effect of the force control gain $G_t$ on this damper term, and a series damper-inerter term that introduces frequency-dependent dissipation that increases with frequency. The parasitic inertance term is scaled by the factor $\frac{1}{(\alpha+1)^2} - \frac{\alpha}{(\alpha+1)^2} \frac{(B_{ref}-B_f)}{B_f}$, indicating that both control gains $G_m$ and $G_t$ have an equal effect on this inerter term.

\medskip
\noindent \underline{Effective Impedance Analysis through Realization:}  
Further understanding of system dynamics can be obtained by studying the effective impedance of an implementation~\cite{colonnese2015rendered,Kenanoglu2022}. Effective impedance definitions decompose the frequency-dependent impedance function into its fundamental components, where the real positive part is associated with the effective damping, while the imaginary part is assigned to the effective spring and effective inertia based on their phase characteristics. The effective impedance analysis of the realization in Table~\ref{fig:allresults}b, after removing  the rendered Voigt model  
$\frac{\alpha}{(\alpha+1)}\,  K_{ref}$--$\frac{\alpha}{(\alpha+1)}\,  B_{ref}$  and the serial coupling filter  $\left(B_f\!-\frac{\alpha }{(\alpha+1)} \, B_{ref}\right)$--$\left(\!K\!-\frac{\alpha}{(\alpha+1)} \, K_{ref}\right)$ pairs, indicates that the  effective damping of the parasitic dynamics can be computed~as \vspace{-3mm}

\begin{equation}\footnotesize
\!\!\!\!c^{SDEA}_{eff_{Voigt}}\!\!\!\!\!\!(\omega) \!\!= \!\! \frac{ \splitfrac{ [ {B_f }^2 \left[ (B_m \!\!+\! G_m )(1\!+\! \alpha) \right] \!\! -\! B_{ref} B_f  \,\alpha (B_m \!\!+\! G_m) \!+}{ J_m  \alpha (B_{ref} K\! \!-\!\! \!B_f K_{ref}) ] w^2\!\! +\!\!K (B_m\!\!\!+\!\!G_m) [ K\!\! +\!\alpha (K \!\!\! -\!\!\!K_{ref} )] } }{{{B_f }^2 (\alpha + 1)^2 }\,w^2 \!+\!K^2 (\alpha+1)^2}
    \label{eqn:SDEAvoigteffecdamp}
\end{equation} \normalsize \vspace{-4mm}

Eqn.~(\ref{eqn:SDEAvoigteffecdamp}) converges to $\sigma (B_m + G_m)$ at low frequencies, while it approaches to $\sigma (B_m + G_m)+ c_{1v}$ at high frequencies. 

Similarly, one can compute the effective inertance of the parasitic dynamics as \vspace{-1mm}
\begin{equation} \footnotesize
b^{SDEA}_{eff_{Voigt}}(\omega)\! =\! \frac{ \splitdfrac{B_f [B_f \!J_m -\alpha\!J_m \!{\left(B_{ref}\!\! -B_f \right)}]\,w^2\!\! +J_m \,K^2 (1 +\alpha)+}{ \alpha(B_f K_{ref} \!-\!\!B_{ref} K )(B_m\!+\! \!G_m )\!\!-\!J_m \,K\,K_{ref} \,\alpha} }{{B_f^2 \left( \alpha+1 \right)^2} \, w^2 +K^2 (\alpha+1)^2}
    \label{eqn:SDEAvoigteffecmass}
\end{equation} \normalsize

At the low frequency range, Equation~(\ref{eqn:SDEAvoigteffecmass}) converges to $b_{1v}+b_{2v}$, whereas at the high frequency range, it approaches to $b_{1v}$. Accordingly, the parasitic damping of  $\sigma  (B_m+G_m)$ affects the Voigt model rendering performance at the low frequency range, while a parasitic inertance of $b_{1v}+b_{2v}$ is also effective in this frequency range. The force controller $G_t$ can effectively mitigate the parasitic damping at low frequencies. The effective parasitic damping increases with frequency and  $c_{1v}$ is added to  $\sigma  (B_m+G_m)$ at the high frequency range.  On the other hand, the effective parasitic inertance decreases with frequency, and $b_{1v}$ becomes more dominant at high frequencies. Hence, the effect of inertance at the low frequency range can be attenuated by both $G_t$ and $G_m$ gains. 
  
Note that, for large controller gains $G_t$ and $G_m$, the dynamics of the parasitic impedance becomes more decoupled from the rendered impedance $ \frac{\alpha}{(\alpha+1)}\,  K_{ref} $--$\frac{\alpha}{(\alpha+1)}\,  B_{ref}$ through the serial coupling filter. Furthermore, the rendered impedance converges to the desired Voigt model.

\medskip
\subsubsection{Series Elastic Actuation~(SEA)}
When the torque and the motion controllers are proportional, the impedance at the interaction port of SEA under VSIC during Voigt rendering is $Z_{Voigt}^{SEA^{P{\text -}P}}(s) =$ \vspace{-3mm}

\begin{equation} \small
\frac{J_m \,K\,s^2 +{\left(B_m \,K+G_m \,K+B_{ref} \,K\,\alpha \right)}\,s+K\,K_{ref} \,\alpha }{J_m \,s^3 +{\left(B_m +G_m \right)}\,s^2 +{\left(K+K\,\alpha \right)}\,s}
    \label{eqn:SEAvoigtPP}
\end{equation} \normalsize \vspace{-4mm}

\medskip
\noindent \underline{Passive Physical Equivalent:} A realization of Eqn.~(\ref{eqn:SEAvoigtPP}) characterizing SEA under VSIC during Voigt model rendering when both controllers are~P is presented in Table~\ref{fig:allresults}e. The parameters of this realization include $c_{2v}~=~\frac{B_m +G_m +B_{ref} \alpha }{\alpha +1}-\frac{\alpha \,K_{ref} \,{\left(B_m +G_m \right)}}{K\,{{\left(\alpha +1\right)}}^2 }$.  The rest of the terms are relatively long and complicated; hence, they are presented as a Matlab script that allows for a numerical means of checking for the non-negativeness of each element\footnote{The Matlab script that presents the parameters of the realization in Table~\ref{fig:allresults}e is available for download at\\ \texttt{\url{https://hmi.sabanciuniv.edu/SEA_Voigt_Realization.m}}.}.

\medskip
\noindent \underline{Haptic Rendering Performance through Realization:} The physical realization of SEA under VSIC during Voigt model rendering in Table~\ref{fig:allresults}e indicates two main branches in parallel: a spring and a branch capturing the dynamics governed by a topology of damper-inertance terms that are coupled to the system through a spring in series. The parallel spring  $\frac{\alpha}{(\alpha+1)}\,K_{ref}$ indicates that, SEA can render the desired spring levels as the dominant behavior of the output impedance function in the low-frequency range. 

Furthermore, in the realization presented in Table~\ref{fig:allresults}e, both damping elements $c_{2v}$ and $c_{3v}$ are functions of $B_{ref}$. Since $c_{2v}$ is more dominant than $c_{3v}$ in the low-frequency range, it can be concluded that $c_{2v}$ mainly contributes to the rendered damping, while the effect of $c_{3v}$ is added to $c_{2v}$ as the frequency increases. 

When a high value of $G_t$ is selected, $c_{2v}$ approaches $B_{ref}$ at low frequencies, indicating that the damping in the system approaches to $B_{ref}$. However, please note that the passivity of the system dictates that $B_{ref}$ should be negative, while feasibility of the realization necessitates that $c_{2v}$ cannot be negative. Accordingly, the realization becomes infeasible before $c_{2v}$ can converge to $B_{ref}$. Similarly, as  $G_t$ increases, the total damping in the system approaches to zero. Finally, $b_{3v}$ acts as a  frequency-dependent parasitic inertence term because of its serial connection with $c_{2v}$.

The coupling filter consists of $K-\frac{\alpha}{(\alpha+1)}\,K_{ref}$, indicating that $c_{2v}$ and $c_{3v}$ become more coupled with the rest for the system with the lower choices of $K_{ref}$. This also implies that pure damping can be rendered at the lower frequency range with the selection of low $K_{ref}$ values.

Please note that the maximum damping that can be passively rendered with SEA under VISC during Voigt model rendering is upper bounded by $\frac{B_m +G_m}{\alpha +1}$ and lower bounded by zero. Hence, the plant damping $B_m$ and the VSIC controller gains $G_t$ and $G_m$ dictate the damping upper bound, while the (negative) $B_{ref}$ acts as a control parameter that adjusts the amount of damping compensation in the system during Voigt model rendering.

\medskip
\noindent \underline{Effective Impedance Analysis through Realization:} To analyze the effective impedance of the realization in Table~\ref{fig:allresults}e, the  effective damping and inertance are computed after removing the rendered virtual stiffness  $\frac{\alpha}{(\alpha+1)}\,  K_{ref} $  and the serial coupling filter $\!K\!-\frac{\alpha}{(\alpha+1)} \, K_{ref}$  from the system. The computed effective damping converges to $c_{2v}$ at low frequencies, while it approaches to $c_{2v}$+$c_{3v}$ at high frequencies. Similarly, the effective inertance converges to $b_{3v}$ at low frequencies, while it approaches zero at high frequencies. Therefore, $c_{2v}$ is the dominant damping in the low-frequency range, and $c_{3v}$ is added to $c_{2v}$ as the frequency increases. While the parasitic effect of $b_{3v}$ exists in low frequencies, this effect will not be dominant in this frequency range, since the branch including $c_{2v}$ is more dominant than serial $c_{3v}$--$b_{3v}$ pair. Due to their serial connection, the effect of $b_{3v}$ becomes higher as the frequency increases, but at the same time, effective inertance goes to zero at high frequencies.

%%%% SPRING RENDERING 
%%%%%%%%%%%%%%%%%%%%%%%%%%%%%%%%%%%%%%%%%%%%%%%%%%%%%%%%%%%%%%%%%%%%%%%%%%%%%%%%%%%%%%%%%%%%%%%%%%%%%%%%%%
%%%%%%%%%%%%%%%%%%%%%%%%%%%%%%%%%%%%%%%%%%%%%%%%%%%%%%%%%%%%%%%%%%%%%%%%%%%%%%%%%%%%%%%%%%%%%%%%%%%%%%%%%%
%%%%%%%%%%%%%%%%%%%%%%%%%%%%%%%%%%%%%%%%%%%%%%%%%%%%%%%%%%%%%%%%%%%%%%%%%%%%%%%%%%%%%%%%%%%%%%%%%%%%%%%%%%
%%%%%%%%%%%%%%%%%%%%%%%%%%%%%%%%%%%%%%%%%%%%%%%%%%%%%%%%%%%%%%%%%%%%%%%%%%%%%%%%%%%%%%%%%%%%%%%%%%%%%%%%%%
%%%%%%%%%%%%%%%%%%%%%%%%%%%%%%%%%%%%%%%%%%%%%%%%%%%%%%%%%%%%%%%%%%%%%%%%%%%%%%%%%%%%%%%%%%%%%%%%%%%%%%%%%%
%%%%%%%%%%%%%%%%%%%%%%%%%%%%%%%%%%%%%%%%%%%%%%%%%%%%%%%%%%%%%%%%%%%%%%%%%%%%%%%%%%%%%%%%%%%%%%%%%%%%%%%%%%

\vspace{-3mm}
\color{black}
\subsection{Spring Rendering}

In this subsection, we present passive physical equivalents of SDEA/SEA during linear spring rendering with their corresponding feasibility conditions.

\smallskip
\subsubsection{Series Damped Elastic Actuation~(SDEA)}

When the torque and the motion controllers are proportional, the impedance at the interaction port of SDEA under VSIC during linear spring rendering equals to
\begin{equation}
\scriptsize
    Z_{spring}^{SDEA^{P{\text -}P}}\!\!\!\!\!\!\!\!(s) \!= \!\frac{\splitdfrac{{B_f \,J_m }\,s^3 +{\left(B_f \,(B_m +G_m) +J_m \,K\right)}\,s^2} {+{\left( K \,(B_m +G_m) +B_f \,K_{ref} \,\alpha \right)}\,s+K\,K_{ref} \,\alpha }}{J_m \,s^3 +{\left(B_f (1+\alpha) +B_m +G_m \right)}\,s^2 +{K(\alpha + 1) }\,s}
    \label{eqn:SDEApsringPP}
\end{equation} \normalsize
%
%\noindent where $\alpha=G_m G_t$.  %Note that when $(\alpha+1)=0$, virtual springs cannot be rendered.
%The passivity of $Z_{spring}^{SDEA^{P {\text -} P}}(s)$ is checked according to Theorem~\ref{theorem:positivereallness}.  

\medskip
\noindent \underline{Passive Physical Equivalent:} A minimal realization of Eqn.~(\ref{eqn:SDEApsringPP}) characterizing SDEA under VSIC during spring rendering when both controllers are P is presented in Table~\ref{fig:allresults}c, where $c_{1s}=\frac{K_{ref} \,\alpha \,{\left(B_f (\,B_m + \,G_m )-J_m \,K\right)}}{B_f \,K\,{{\left(\alpha +1\right)}}^2 }$, $b_{1s}=\frac{K_{ref} \,\alpha \,{\left(B_f (\,B_m + \,G_m ) -J_m \,K\right)}}{K^2 \,{{\left(\alpha +1\right)}}^2 }$, and $\sigma=\frac{1}{\alpha+1} - \frac{\alpha}{(\alpha+1)^2} \frac{K_{ref}}{K}$.

For the realization in Table~\ref{fig:allresults}c to be feasible, all physical components of the model should be non-negative. Hence, $(\alpha+1)$ should be positive. Furthermore, the non-negativeness of the coupling spring imposes $\frac{\alpha}{\alpha+1} \, K_{ref} \le K$. The non-negativeness $\sigma (B_m+G_m)$ is guaranteed if $(B_m+G_m) > 0$ and $\frac{\alpha}{\alpha+1} \, K_{ref} \le K$ are simultaneously satisfied. %As $(\alpha+1)$ goes to zero, $K-\frac{\alpha }{\alpha +1}  \,K_{ref}$ converges to $-\infty$. 
 The virtual stiffness is rendered as $\frac{\alpha }{\alpha +1}  \,K_{ref}$. % approaches zero, the outer spring which represents the virtual stiffness to be rendered, converges to zero. 
  The conditions for the non-negativeness of $c_{1s}$ and $b_{1s}$ can be derived as  \vspace{-3mm}

\begin{equation}
   J_m \, \frac{  K}{B_f} \le (\,B_m + \,G_m ) 
    \label{eqn:SDEAspringPP_k1s}
\end{equation} \vspace{-4.5mm}

\noindent which indicates $(B_m+G_m) > 0$.

\medskip
\noindent \underline{Haptic Rendering Performance through Realization:} 
The physical realization of SDEA during linear spring rendering in Table~\ref{fig:allresults}h indicates two branches in parallel: an ideal spring $\frac{\alpha}{(\alpha+1)} \, K_{ref}$ whose stiffness approaches to  $K_{ref}$ as the controller gains get large and parasitic dynamics governed by damper-inertance elements that are serially coupled to the system with a spring-damper pair $\left(K-\frac{\alpha}{(\alpha+1)} \,  K_{ref} \right)-B_f$ in parallel. %The main differences are due to the physical filter damping $B_f$ appearing in parallel to the coupling spring  $K-\frac{\alpha}{(\alpha+1)} \,  K_{ref}$ and the more complicated form of the parasitic damper-inertance dynamics. 

Due to the existence of the physical filter damping $B_f$ in parallel to the coupling spring, the parasitic dynamics cannot be completely decoupled from the system as the controller gains $G_t$ and $G_m$ increase, since $B_f$ term dominates the coupling at the intermediate and high frequencies.  Table~\ref{fig:allresults}h indicates that the parasitic effects of the damper $\sigma(B_m+G_m)$ and the inerter $J_m/(\alpha+1)$  terms decrease with the choice of high controller gains. In particular, $G_t$ has a more dominant effect on the damper term, while $G_m$ and $G_t$ gains affect the inerter term in the same manner, as they are multiplicative. In addition to these parallel damper-inerter terms, SDEA realization includes frequency-dependent dissipative effect which consists of serial damper-inerter terms. %When $K_{ref} =0$, the parasitic dynamics converge to that of null impedance rendering. 

\medskip
\underline{Effective Impedance Analysis through Realization:} 
An effective impedance analysis of the parasitic dynamics of the realization in Table~\ref{fig:allresults}c indicates that the  effective damping of Eqn.~(\ref{eqn:SDEApsringPP}) after removing the serial coupling filter  $B_f$-$\left(K-\frac{\alpha}{(\alpha+1)} \, K_{ref}\right)$ pair and the rendered stiffness  $\frac{\alpha}{(\alpha+1) }\, K_{ref}$ can be computed as  \vspace{-3.5mm}

\begin{equation} \small
c^{SDEA}_{eff_{PP}} = \frac{ \splitdfrac{{B_f \left({B_f } \,(B_m +\,G_m )(\alpha + 1) -J_m \,K_{ref} \,\alpha \right)}\,\omega^2 +}{K \, ( K (B_m +G_m ) (\alpha + 1 ) -\,K_{ref} \,\alpha (B_m + G_m ) )}}{{{B_f }^2 \left( \,\alpha +1 \right)^2}\,\omega^2 +K^2 (\,\alpha + 1 )^2}
    \label{eqn:SDEAPPspringeffecdamp}
\end{equation} \normalsize  \vspace{-4mm}

At low frequencies, Eqn.~(\ref{eqn:SDEAPPspringeffecdamp}) converges to $\sigma (B_m+G_m)$, while at high frequencies,  Eqn.~(\ref{eqn:SDEAPPspringeffecdamp}) approaches to $\sigma (B_m+G_m) + c_{1s}$. Similarly, the effective inertence for the parasitic dynamics of Eqn.~(\ref{eqn:SDEApsringPP}) can be computed as  \vspace{-4mm}

\begin{equation} \small
\!\! b^{SDEA}_{eff_{PP}} \!\! =  \!\!\frac{\splitdfrac{{\left({B_f }^2 \,J_m (\alpha+1) \right)}\,\omega^2+}{ J_m \,K^2 (\alpha+1) +B_f\,K_{ref} \,\alpha (B_m + G_m) -J_m \,K\,K_{ref} \,\alpha }}{{{B_f }^2 \left( \,\alpha +1 \right)^2}\,\omega^2 +K^2 (\,\alpha + 1 )^2}
    \label{eqn:SDEAPPspringeffecmass}
\end{equation} \normalsize  \vspace{-4mm}

At low frequencies, Eqn.~(\ref{eqn:SDEAPPspringeffecmass}) converges to $\frac{J_m}{\alpha+1}+b_{1s}$, while at high frequencies,  Eqn.~(\ref{eqn:SDEAPPspringeffecmass}) approaches to $\frac{J_m}{\alpha+1}$.

\bigskip
\color{black}
\subsubsection{Series Elastic Actuation~(SEA)}

When both the motion and torque controllers are proportional, the impedance at the interaction port of SEA under VSIC during spring rendering equals to
\begin{equation} \small
 \!\!\!\!   Z_{spring}^{SEA^{P {\text -} P}}(s) \!\! = \frac{{J_m \,K}\,s^2 +{\left(B_m +G_m \right)} \,K \,s+\alpha \,K\,K_{ref} }{J_m \,s^3 +{\left(B_m +G_m \right)}\,s^2 +{\left(\alpha + 1 \right) K}\,s}
    \label{eqn:SEApsringPP}
\end{equation} \normalsize
%
%The passivity of $Z_{spring}^{SEA^{P {\text -} P}}(s)$ is checked according to Theorem~\ref{theorem:positivereallness}.  

%\medskip
\noindent \underline{Passive Physical Equivalent:} A minimal realization of Eqn.~(\ref{eqn:SEApsringPP}) characterizing SEA under VSIC during spring rendering when both controllers are~P is presented in Table~\ref{fig:allresults}f, where $\sigma=\frac{1}{\alpha+1} - \frac{\alpha}{(\alpha+1)^2} \frac{K_{ref}}{K}$.

For the realization in Table~\ref{fig:allresults}f to be physically feasible, all of the components of the model should be non-negative. All components in the realization are guaranteed to be non-negative, if $K \ge \frac{\alpha }{\alpha +1}  \,K_{ref}$ is satisfied, and $(B_m + G_m)$,  $\frac{\alpha }{\alpha +1}  \,K_{ref}$, and $(\alpha+1)$ are positive. %As $(\alpha+1)$ goes to zero, $K-\frac{\alpha }{\alpha +1}  \,K_{ref}$ converges to $-\infty$. Furthermore, 
%As $\frac{\alpha }{\alpha +1}$ approaches to zero, the outer spring which represents the virtual stiffness to be rendered, converges to zero. %\fbox{WHAT IS WRONG WITH THAT in terms of realization?} 

\medskip

\noindent \underline{Haptic Rendering Performance through Realization:} The physical realization of SEA during linear spring rendering in Table~\ref{fig:allresults}f  indicates two branches in parallel: an ideal spring $\frac{\alpha }{(\alpha+1)}\, K_{ref}$ whose stiffness approaches to $K_{ref}$ as the controller gains get large and parasitic dynamics governed by a damper-inertance pair in parallel that is coupled to the system with a spring in series. The stiffness of the coupling spring is given by $K-\frac{\alpha }{(\alpha+1)} \, K_{ref}$; hence, the parasitic dynamics get more decoupled from the system as the controller gains $G_t$ and $G_m$ increase. Note that, since the coupling spring needs to be positive for feasibility, this spring imposes an upper bound on $K_{ref}$ that can be passively rendered. The parasitic damper-inertance dynamics is scaled by $\sigma=\frac{1}{\alpha+1} - \frac{\alpha}{(\alpha+1)^2} \frac{K_{ref}}{K}$, indicating that $G_t$ has a significant effect for damper term, while both $G_m$ and $G_t$ equally affect the inerter term. Furthermore, the parasitic dynamics decrease with the choice of higher $K_{ref}$ values. When $K_{ref} =0$, the parasitic dynamics converge to that of null impedance rendering. 

\medskip
\noindent \underline{Effective Impedance Analysis through Realization:}
The effective impedance of the system dynamics after the serial physical filter $K-\frac{\alpha }{(\alpha+1)} \, K_{ref}$ and rendered stiffness $\frac{\alpha}{(\alpha+1)}\, K_{ref}$ are extracted, is dominated by the damper term $\sigma (B_m+G_m)$ in the low frequency range. Therefore, the spring rendering performance can be increased in the low frequency range by attenuating the effects of this damper term. Similarly, the high frequency behavior of these parasitic dynamics is dictated by the inerter term $\sigma J_m$. %Hence, the passive equivalents provide an explicit representation for the effective impedance analysis. 

\vspace{-1mm}
\begin{remark}[2]
Table~\ref{fig:allresults}  indicates that there exists continuity among the realizations of both SDEA and SEA under VSIC; that is,  by setting $B_{ref}\!=\!0$ in the realization of Voigt model rendering, the realization of spring rendering can be achieved. Similarly, the realization of null impedance rendering can be recovered from the realization of Voigt model rendering by setting $B_{ref} \!=\!0$ and $K_{ref} \!=\! 0$, simultaneously. %Hence, haptic rendering performance and effective impedance analysis for the spring and null impedance can be directly studied through the Voigt model rendering results.
\end{remark}

\vspace{-2mm}

\begin{remark}[3]
Table~\ref{fig:allresults} also indicates that there exists continuity among the realizations of SDEA under VSIC and SEA under VSIC; that is, by setting $B_f\!=\!0$ in the SDEA realizations, the realizations of SEA can be recovered. Note that since the Voigt model rendering realization in Table~\ref{fig:allresults}b is valid only for the positive values of $B_{ref}$, while SEA under VSIC is not passive for $B_{ref} >0$, no such realization exists for SEA. On the other hand, the realizations in Table~\ref{fig:allresults}a and Table~\ref{fig:allresults}e are both valid for  $B_{ref} <0$ and display the desired continuity as their passive parameter ranges overlap.
\end{remark}

\vspace{-4mm}
\section{Passive Physical Equivalents vs Passivity} \label{Sec:Passivity}

In this section,  we present the necessary and sufficient conditions for the passivity of SDEA and SEA under VSIC with proportional~(P)  controllers while rendering Voigt and linear models, without imposing a non-negativity assumption on the controller gains. Please note that the inner motion control loop is considered to be asymptotically stable throughout the analyses, imposing $(B_m + G_m) > 0$. We also compare feasibility conditions of passive physical equivalents with the necessary and sufficient conditions for the passivity.

Proposition~\ref{theorem:SDEAPPvoigttheorem} presents the necessary and sufficient conditions for the passivity of SDEA under VSIC while rendering Voigt models which is presented in Eqn. (\ref{eqn:SDEAvoigtPP}). %\vspace{-1mm} 
%,  when $G_t$ and $G_m$ consist of proportional gains and the inner motion control loop is stable

\vspace{-1mm}
\begin{proposition}
Consider Voigt model rendering with SDEA under VSIC as in Figure~\ref{fig:VSICclosed}, where the torque and velocity controllers consist of proportional gains $G_t$ and $G_m$, respectively. Let the physical plant parameters be positive, while the controller gains are allowed to be negative as long as the inner motion control loop is asymptotically stable. Then, the following inequalities serve as the necessary and sufficient conditions for establishing the passivity of $Z_{Voigt}^{SDEA^{P{\text -}P}}\!(s)$: %\vspace{-2mm}

\begin{enumerate}
\item[(i)] $0 < \frac{\alpha}{\alpha +1} \, K_{ref} \le \left( 1+\frac{\alpha \, B_{ref} }{B_m +G_m} \right) \, K $, \textrm{and}
%$K\, \ge K_{ref} \, \frac{\alpha}{(\alpha+1)} \, \frac{B_m+G_m}{ B_m + G_m +   B_{ref} \, \alpha}$, \textrm{and}
\item[(ii)]  $-(B_m+G_m) \le \alpha \, B_{ref}$,  \textrm{and}
%\item[(iv)]  $0 < \frac{\alpha}{\alpha+1} K_{ref} $,  \textrm{and}
\item[(iii)]  $0 < (\alpha+1) $, \textrm{and}
\item[(iv)]  $0 < (B_m+G_m) $, \textrm{and}
\item[(v)] \! \! \small $\!\!-2 J_m  \!  \sqrt{B_f K \left[ (B_m \!\!+ \!\!G_m\! \!+ \!\! B_{ref} \alpha ) \! K {(\alpha \! + \!1)} \! - \! \!(B_m\! + \! \!G_m)  K_{ref} \alpha \right]} \nonumber \\
\!\! \le B_f \,{\left(B_m \!+\!\! G_m \!+ B_{ref} \, \alpha \right)}\, \left[ B_m\! + G_m\! +\!\! B_f \, (\alpha+\!\!1) \right] \nonumber \\
\!\!\!\! -\!(B_{ref} \,K \!+ B_f \, K_{ref}) \, J_m \, \alpha$.
\normalsize
\end{enumerate}
\label{theorem:SDEAPPvoigttheorem}
\end{proposition}
\vspace{-1mm}

The proof is presented in Appendix~A.

 \vspace{-1mm}
\begin{remark}[1]
A (more conservative) set of sufficient conditions can be derived by considering  Conditions~(i)--(iv) together with the following inequality instead of Condition~(v):  \vspace{-2.mm}

 \footnotesize
\begin{equation}  %\vspace{-3mm}
J_m \le \frac{{ B_f \, \left(B_m \! +\!G_m \!+\! B_{ref} \, \alpha \right)}\,{\left[B_m \! +\! G_m \!+\!B_f \, (1\!+ \!\alpha) \right]}}{\left(B_f \, K_{ref}\! +\! B_{ref} \,K \right) \, \alpha } 
    \label{eqn:SDEAvoigtd4}
\end{equation} \normalsize
\end{remark}

\smallskip
\noindent \ul{Feasibility of Passive Realization vs Passivity for Voigt Model Rendering with SDEA:}
The feasibility conditions in Eqns.~(\ref{eqn:SDEAspringPP_spring})--(\ref{eqn:SDEAspringPP_extraterm}) serve as a set of sufficient conditions for the passivity of SDEA under VISC. In particular, Eqn.~(\ref{eqn:SDEAspringPP_spring}) is a more conservative condition than Condition~(i) of Proposition~\ref{theorem:SDEAPPvoigttheorem} as shown below: \vspace{-3mm}

 \small
\begin{equation}
 K\, \ge K_{ref} \, \frac{\alpha}{(\alpha+1)} \ge K_{ref} \, \frac{\alpha}{(\alpha+1)} \, \frac{B_m+G_m}{ B_m + G_m +   B_{ref} \, \alpha}
\end{equation} \normalsize 

Similarly, Eqn.~(\ref{eqn:SDEAspringPP_extraterm})  imposes a constraint that is more conservative than the sufficiency condition in Eqn.~(\ref{eqn:SDEAvoigtd4}), as can be shown by substituting Condition~(i) of Proposition~\ref{theorem:SDEAPPvoigttheorem} into Eqn.~(\ref{eqn:SDEAvoigtd4}) and noting that the plant parameters are positive.  %\vspace{-3mm}

 \small
\begin{eqnarray} \vspace{-6mm}
 \!\! J_m \!\!\!\! &\le& \!\!\!\!  \frac{B_f}{K} \,  (B_m\! +G_m)   \nonumber \\
     \!\!\!\! &\le& \!\!\!\! \frac{{B_f \, \left(B_m \! +G_m \!+ B_{ref} \, \alpha \right)}\,{\left[B_m \! + G_m \!+B_f \, (1+ \alpha) \right]}}{\left(B_f \, K_{ref} + B_{ref} \,K \right) \, \alpha } 
\end{eqnarray} \normalsize

Consequently, the feasibility of the realization in Table~\ref{fig:allresults}b provides a set of sufficient conditions for the passivity of Eqn.~(\ref{eqn:SDEAvoigtPP}); the realization in Table~\ref{fig:allresults}b is valid when Eqns. (\ref{eqn:SDEAspringPP_spring})--(\ref{eqn:SDEAspringPP_extraterm}) are satisfied with positive $\frac{\alpha}{\alpha+1 }B_{ref}$ and $\frac{\alpha}{\alpha+1 } K_{ref}$ values.

\medskip 

Corollary~\ref{theorem:SEAPPvoigttheorem} presents the necessary and sufficient conditions for the passivity of SEA under VSIC while rendering Voigt models as in Eqn. (\ref{eqn:SEAvoigtPP}).  Corollary~\ref{theorem:SEAPPvoigttheorem} shows that SEA can render Voigt model if controllers are allowed to be negative.%\vspace{-1.5mm}

\begin{corollary}
Consider Voigt model rendering with SEA under VSIC as in Figure~\ref{fig:VSICclosed} with $B_f$=0, where  the torque and velocity controllers consist of proportional gains $G_t$ and $G_m$, respectively.
Let the plant parameters be positive, while the controller gains are allowed to be negative as long as the inner motion control loop is asymptotically stable. Then, the following inequalities serve as the necessary and sufficient conditions for establishing the passivity of $Z_{Voigt}^{SEA^{P{\text -}P}}\!(s)$: %\vspace{-2mm}

\begin{enumerate} 
\item[(i)] $ 0 < \frac{\alpha }{\alpha +1} \, K_{ref} \le \left(1+\frac{\alpha \, B_{ref} }{B_m +G_m } \right) \, K $, \textrm{and}
\item[(ii)] $- \, (B_m + G_m) \le \alpha \, B_{ref} \,  \le \, 0$, \textrm{and}
%\item[(iii)]  $\frac{\alpha}{\alpha+1} K_{ref} > 0$,  \textrm{and}
\item[(iii)]  $0 < (\alpha + 1) $,  \textrm{and}
\item[(iv)]  $0 < (B_m + G_m)  $.
\end{enumerate}
\label{theorem:SEAPPvoigttheorem}
\end{corollary}
\vspace{-1mm}
\noindent The proof follows from Proposition~\ref{theorem:SDEAPPvoigttheorem} by substituting $B_f \! = \!0$.
Corollary~\ref{theorem:SEAPPvoigttheorem} necessities $B_{ref}$ and $\alpha$  have opposite signs. 
%When $\alpha \ge 0$, lower $B_{ref}$ gains make the upper bound on $K_{ref}$ more conservative. When $\alpha \le 0$, $B_{ref}$ should be selected as positive, but because of the other conditions gains will be deficient for acceptable rendering performance. 

\medskip
\noindent \ul{Feasibility of Passive Realization vs Passivity for Voigt Model Rendering with SEA:} If we consider $G_m$ and $G_t$ to be non-negative, then symbolic substitutions and numerical evaluations indicate that the non-negativeness of $c_{2v}$ imposes Condition~(i) of Corollary~\ref{theorem:SEAPPvoigttheorem}. Moreover, if we substitute the non-negativeness condition of $c_{2v}$ into $b_{3v}$ and $c_{3v}$, we observe that $B_{ref}$ should be negative for non-negativeness of $b_{3v}$ and $c_{3v}$. Hence, the feasibility of the realization in Table~\ref{fig:allresults}e provides sufficient conditions for the passivity of the system. Accordingly, if we consider that the controller gains are positive, then the realization in Table~\ref{fig:allresults}e is valid as long as $B_{ref}$ is negative, and Condition~(i) of Corollary~\ref{theorem:SEAPPvoigttheorem} are satisfied with non-negative $b_{3v}$ and $c_{3v}$ values. 

The Voigt model rendering realization for SEA presented in Table~\ref{fig:allresults}e is valid only for the negative values of $B_{ref}$, as positive values of $B_{ref}$ do not result in passive Voigt model rendering for SEA under VSIC with P-P controllers. The realization in Table~\ref{fig:allresults}e can be recovered from the SDEA realization in Table~\ref{fig:allresults}a, when $B_f$ is set to zero. On the other hand, the realizations in Table~\ref{fig:allresults}b and~Table~\ref{fig:allresults}e have distinct topology as they cover non-overlapping system parameters.

\begin{remark}[2]
If $B_{ref}$ is set to zero, Eqn.~(\ref{eqn:SDEAvoigtPP}) reduces to spring rendering with SDEA under VSIC as in Eqn.~(\ref{eqn:SDEApsringPP}). Similarly, if $B_{ref}$ and $K_{ref}$ are set to zero, Eqn. (\ref{eqn:SDEAvoigtPP}) reduces to null impedance rendering with SDEA under VSIC. Hence, the necessary and sufficient conditions for spring and null impedance rendering can be derived from Proposition~\ref{theorem:SDEAPPvoigttheorem}; the proofs follow by substituting~$B_{ref}\! = \! 0$ and~$B_{ref}\! = K_{ref}\! =\! 0$. %\vspace{1mm}
%of Corollary~\ref{theorem:SDEAPPspringtheorem} is presented in the Supplementary Document~\cite{supplementary}.
\end{remark}

\noindent \ul{Feasibility of Passive Realization vs Passivity for Spring Rendering with SDEA:}  
The feasibility conditions for the realization in Table~\ref{fig:allresults}c provide sufficient conditions for the passivity of Eqn.~(\ref{eqn:SDEApsringPP}). This can be shown by first considering a sufficient condition for the passivity that is ensured by imposing a non-negative value to the intermediate coefficient of the test polynomial as follows \vspace{-1.mm}

\begin{equation}\small
    J_m \le \frac{{\left(B_m +G_m \right)}\,{\left( B_m +G_m + B_f (\alpha+1)  \right)}}{\alpha \,K_{ref} }
    \label{eqn:SDEAspringnonnegatived4}
\end{equation} \normalsize \vspace{-3.7mm}

\noindent Note that replacing the condition provided in Condition~(v) of Proposition~\ref{theorem:SDEAPPvoigttheorem} when $B_{ref}=0$ with the non-negativeness of the intermediate coefficient of the test polynomial provides a (more conservative) sufficient condition for the passivity. This condition still needs to be considered together with the other necessary conditions of the non-negativeness of the highest and lowest coefficients of the test polynomial. Eqns.~(\ref{eqn:SDEAspringPP_k1s}) and~(\ref{eqn:SDEAspringnonnegatived4}) can be arranged together as \vspace{-3.mm}

\begin{equation} \small
\small
    J_m \le  \frac{ (\,B_m + \,G_m ) \, B_f}{K}  \le \frac{{\left(B_m +G_m \right)}\,{\left(B_m +G_m +B_f (\alpha +1) \right)}}{\alpha\,K_{ref} }
    \label{eqn:SDEAspringd4k1scomb}
\end{equation} \normalsize

  \begin{figure*}[b] 
  \centering \vspace{-3mm}
 \begin{tabular}{ccc}
 \begin{subfigure}[a]{0.315\linewidth}
   \includegraphics[keepaspectratio=true, width=\linewidth]{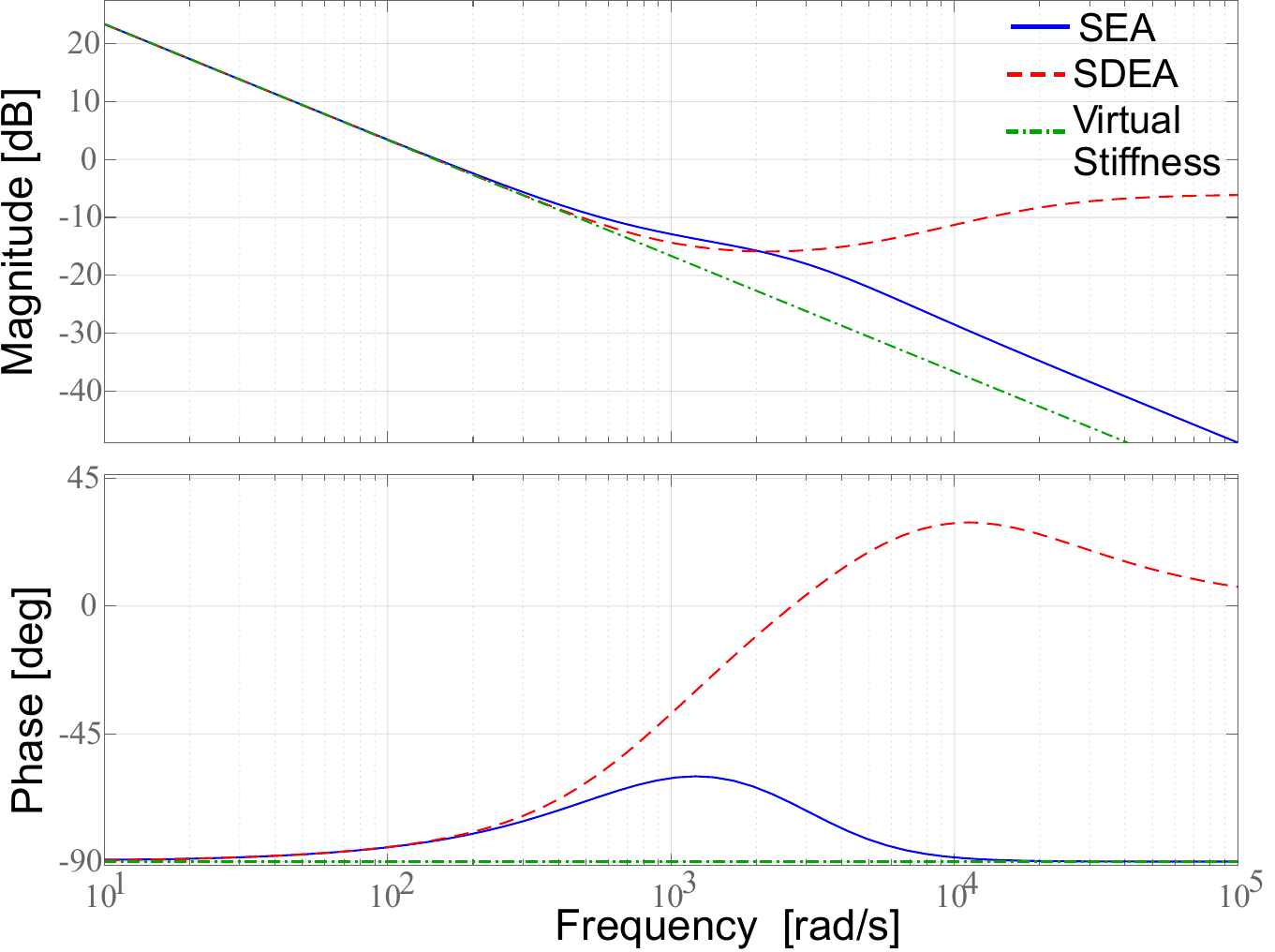}
              \vspace{-6mm}
    \caption{Linear spring rendering }
    \label{fig:SEAvsSDEAspring}
    \end{subfigure} & 
         \begin{subfigure}[a]{0.315\linewidth}
     \includegraphics[keepaspectratio=true, width=\linewidth]{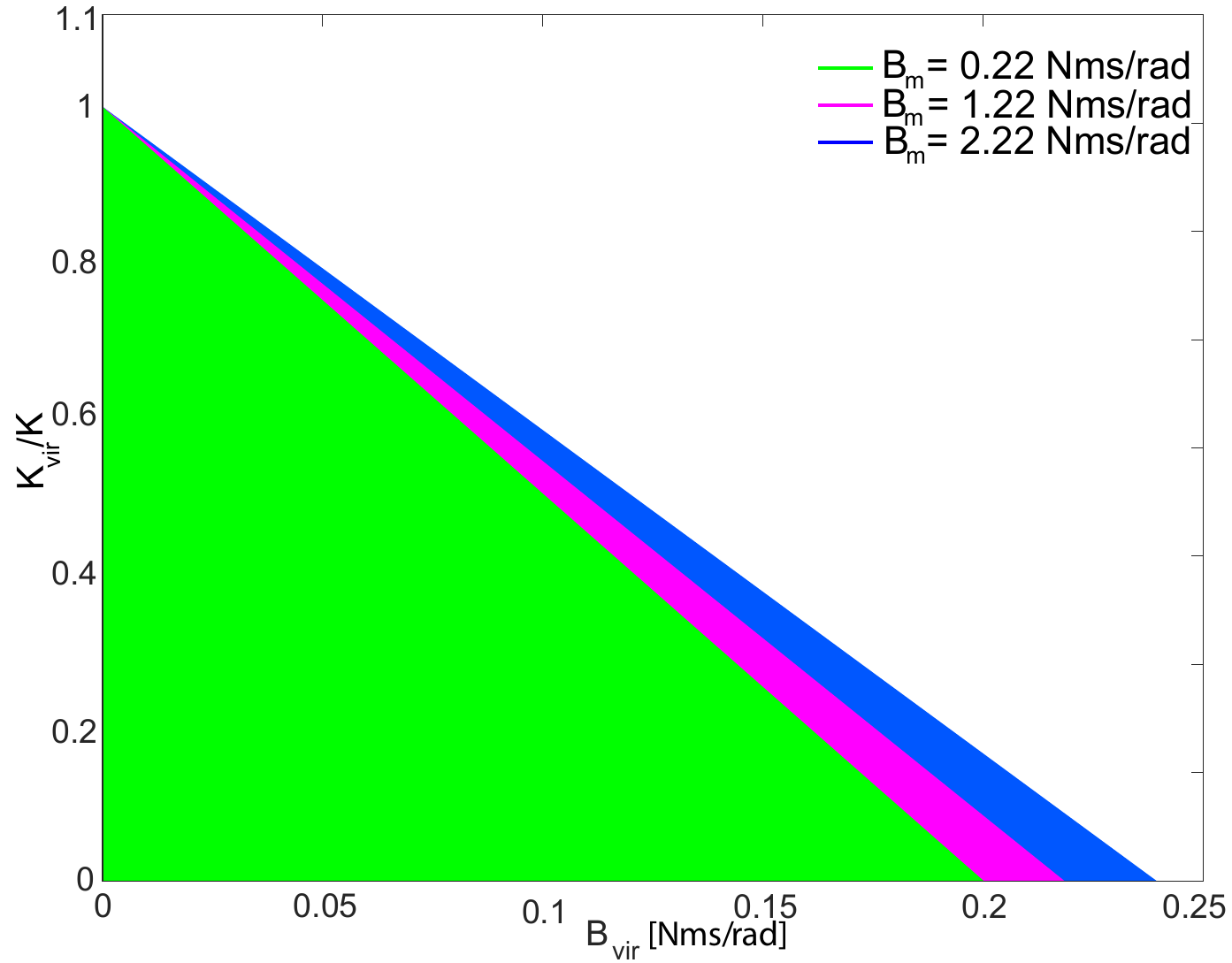}
              \vspace{-6mm}
    \caption{ K-B plot for SEA under VSIC }
    \label{fig:KBplot}
    \end{subfigure}&
    
     \begin{subfigure}[a]{0.315\linewidth}
     \includegraphics[keepaspectratio=true, width=\linewidth]{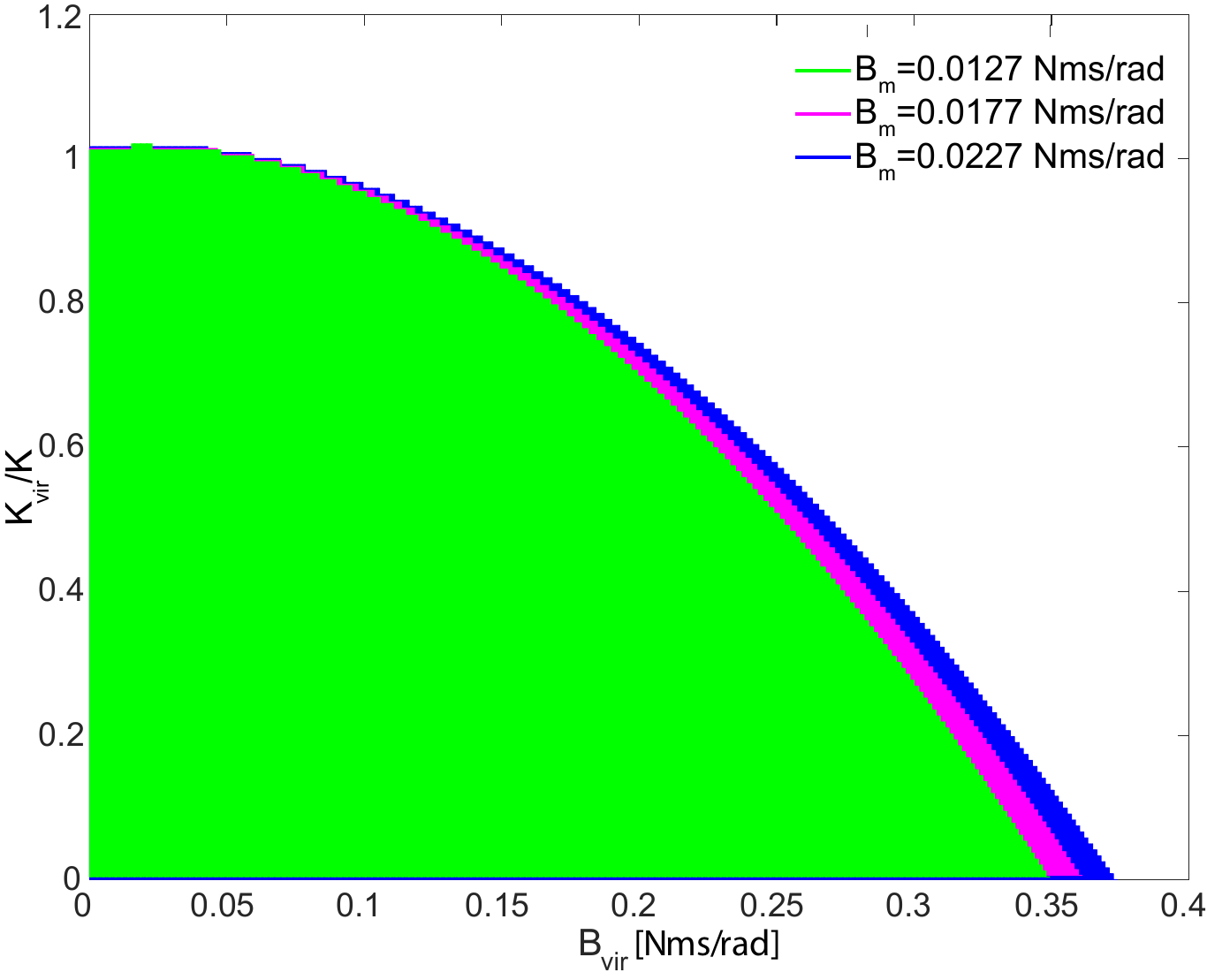}
              \vspace{-6mm}
    \caption{K-B plot for SDEA under VSIC }
    \label{fig:KBplotSDEA}
    \end{subfigure}  

\end{tabular} 
  \vspace{-.6\baselineskip}
    \caption{Spring rendering performance and K-B plot comparison during Voigt model rendering between S(D)EA under VSIC}
        \vspace{-1.25\baselineskip}
    \label{fig:SEAvsSDEArendering}
    %\vspace{-2mm}
  \end{figure*}

\noindent Given Condition (i) of Proposition~\ref{theorem:SDEAPPvoigttheorem} when $B_{ref}=0$ as necessitated by the feasibility of the realization in Table~\ref{fig:allresults}c and the passivity of Eqn.~(\ref{eqn:SDEApsringPP}), 
this inequality is always satisfied. Therefore, Eqn.~(\ref{eqn:SDEAspringPP_k1s}) is a more conservative sufficient condition than the one provided in Eqn.~(\ref{eqn:SDEAspringnonnegatived4}) and 
when Eqn.~(\ref{eqn:SDEAspringPP_k1s}) is satisfied, Condition (v) of Proposition~\ref{theorem:SDEAPPvoigttheorem} when $B_{ref}=0$ is guaranteed to hold. Accordingly, the realization in Table~\ref{fig:allresults}c is feasible and valid, and the sufficient conditions for the passivity of Eqn.~(\ref{eqn:SDEApsringPP}) are satisfied if $(B_m+G_m)$, $(\alpha+1)$, and $\frac{\alpha }{\alpha +1} \, K_{ref}$ are positive, and Eqn.~(\ref{eqn:SDEAspringPP_k1s}) is satisfied. If Condition (v) of Proposition~\ref{theorem:SDEAPPvoigttheorem} when $B_{ref}=0$ is replaced with Eqn.~(\ref{eqn:SDEAspringPP_k1s}), then the necessary and sufficient conditions for the passivity of Eqn.~(\ref{eqn:SDEApsringPP}) can be recovered.

\medskip
\noindent \ul{Feasibility of Passive Realization vs Passivity for Null Impedance Rendering with SDEA:}~The~conditions for the feasibility of the realization in Table~\ref{fig:allresults}d are equivalent to the necessary and sufficient conditions for the passivity of null impedance rendering: if $(B_m+G_m) > 0$ and $(\alpha + 1) > 0 $, then null impedance rendering is passive and all components in Table~\ref{fig:allresults}d are non-negative. Accordingly, the realization is valid as long as the system is passive.

\begin{remark}[3]
If $B_{f}$ and $B_{ref}$ are set to zero, Eqn. (\ref{eqn:SDEAvoigtPP}) reduces to spring rendering with SEA under VSIC as in Eqn. (\ref{eqn:SEApsringPP}). Similarly, if $B_f$, $B_{ref}$, and $K_{ref}$ are set to zero, Eqn. (\ref{eqn:SDEAvoigtPP}) reduces to null impedance rendering with SEA under VSIC. Hence, the necessary and sufficient conditions for spring and null impedance rendering can be derived from Proposition~\ref{theorem:SDEAPPvoigttheorem}. The proof follows from Proposition~\ref{theorem:SDEAPPvoigttheorem} after substituting~$B_{f}\! =\!B_{ref}\! = \! 0$ and~$B_{f}\! =\!B_{ref}\! = K_{ref}\! =\! 0$.
\end{remark}

\medskip
\noindent \ul{Feasibility of Passive Realization vs Passivity for Spring Rendering with SEA:}~The~conditions for the feasibility of the realization in Table~\ref{fig:allresults}f are equivalent to the necessary and sufficient conditions for the passivity of Eqn.~(\ref{eqn:SEApsringPP}): if $K \ge \frac{\alpha }{\alpha +1}  \,K_{ref}$ is satisfied and $(B_m + G_m)$, $(\alpha+1)$, and $\frac{\alpha }{\alpha +1}  \,K_{ref}$ are positive, then Eqn.~(\ref{eqn:SEApsringPP}) is passive and all components in Table~\ref{fig:allresults}f are non-negative. Accordingly, the realization is valid as long as the system is passive.

Feasibility of passive realization vs passivity analysis for null impedance rendering with SEA can be achieved by substituting $K_{ref}=0$ as presented in~\cite{supplementary}.

%xxxxxxx
\section{Performance Comparisons via Realizations} \label{Sec:Codesign}

%however, a fair comparison of the effects of plants and/or controllers with different dynamics on system performance remains a challenging problem. Yusuf~\etal~\cite{Yusuf_multi_2020} proposed performing numerical multi-criteria performance optimization of each closed-loop rendering system and comparison of the non-dominated solution lying on the Pareto-front of each optimization to establish a fair comparison between controllers of different types. While this approach is sound and effective, the numerical nature of the technique limits the generality of conclusions that can be drawn from such comparisons.

This section provides a rigorously symbolic comparison of different plant dynamics on system performance through passive physical equivalents with continuity. In particular, we demonstrate how passive mechanical equivalents enable fair comparisons of SEA and SDEA plant dynamics on the haptic rendering performance. Unlike the case in numerical studies~\cite{Yusuf_multi_2020}, comparisons of closed-loop system dynamics through passive physical equivalents are informative in that their conclusions can be generalized, allowing the designer to make informed decisions among various plants/controllers.  %These comparisons highlight the impact of different plant and controller terms on the closed-loop rendering performance. Furthermore, since there exists continuity among realizations, the effect of each controller term on plant dynamics can be rigorously studied. Moreover, these comparisons are symbolic in nature and do not require  

The main difference between SDEA and SEA plants while rendering Voigt model is that SEA can only compensate for the system damping via negative $B_{ref}$ values as in in Table~\ref{fig:allresults}e, while SDEA can assume both negative and positive $B_{ref}$ values, as in Tables~\ref{fig:allresults}a and \ref{fig:allresults}b to render damping levels lower and higher than the system damping.

For a direct comparison during Voigt model rendering, negative $B_{ref}$ values are considered such that both realizations are feasible for overlapping parameter ranges.
Tables~\ref{fig:allresults}a and \ref{fig:allresults}e present passive mechanical realizations of SDEA and SEA plants, respectively, while rendering Voigt models.
 At low frequencies, the effective stiffness of both realizations approaches the desired output impedance of  $\frac{\alpha}{\alpha+1}K_{ref}$. Similarly, the damping of SDEA in Table \ref{fig:allresults}a converges to $c_{1va}$+$B_f$, while the damping of SEA in Table \ref{fig:allresults}e converges to $c_{2v}$. A closer inspection of the damping values reveals that $c_{1va}$+$B_f$ and $c_{2v}$ are equal to each other, and  both $c_{1va}$ and $c_{2v}$ are modifiable through different values of $B_{ref}$. Both realizations include frequency-dependent dissipation that increases with the frequency. At high frequencies, SDEA and SEA converge to the characteristic of their respective compliant elements. %: $K-B_f$ and $K$.

  %If Tables \ref{fig:allresults}a and \ref{fig:allresults}e damping behavior are compared in the low-frequency range, Table \ref{fig:allresults}a converges to $c_{1va}$+$B_f$, while Table \ref{fig:allresults}e converges to $c_{2v}$. When $c_{1va}$+$B_f$ and $c_{2v}$ are compared, we can see that they are identical. Please note that both $c_{1va}$ and $c_{2v}$ are modifiable with different $B_{ref}$. 
 % Tables \ref{fig:allresults}a and \ref{fig:allresults}e indicate the frequency-dependent dissipation, which contributes to the increase in system damping as the frequency rises.

Tables~\ref{fig:allresults}c and \ref{fig:allresults}f present passive mechanical realizations of SDEA and SEA plants, respectively, while rendering linear springs.
At low frequencies, the effective stiffness of both realizations approaches the desired output impedance of  $\frac{\alpha}{\alpha+1}K_{ref}$. The parasitic terms are more strongly coupled to the desired spring in SDEA, as $B_f$ acts in parallel to the coupling spring $K-\frac{\alpha}{\alpha+1}K_{ref}$. Furthermore, the parasitic inertia term $\frac{J_m}{\alpha+1}$ of SDEA in Table \ref{fig:allresults}c is always greater than the parasitic inertia term $\sigma J_m$ of SEA in Table \ref{fig:allresults}f. Finally, the parasitic damping of SDEA includes an extra frequency-dependent dissipation that increases with the frequency.

Tables~\ref{fig:allresults}d and \ref{fig:allresults}g present passive mechanical realizations of SDEA and SEA plants, respectively, while rendering the null impedance.
The parasitic terms are more strongly coupled in SDEA, as $B_f$ acts in parallel to the coupling spring $K$, while there exists no difference between the parasitic terms. %of SDEA and SEA.

Table~\ref{table:SEAparamters} presents the physical parameters of the S(D)EA plant used in numerical simulations to evaluate the system performance. The proportional controller gains are set as $G_m$~=~10~N-m~s/rad and $G_t$~=~5~rad/(s~N-m), respectively. 

\vspace{-1mm}
\begin{table}[h] \small
\centering %\vspace{-2mm}
\caption{S(D)EA plant parameters used for the simulations}
\vspace{-2mm}
\label{table:SEAparamters}
\resizebox{1\linewidth}{!}{
\begin{tabular}{|c|c|c|c|c|}
\hline
Parameter & $J_m$ & $B_m$ & $K$ & $B_f$ \\ \hline
Value & 0.002~kg-m$^2$ & 1.22~N-m~s/rad & 360~N-m/rad & 0.5~N-m~s/rad \\ \hline
\end{tabular}}
\vspace{-.750\baselineskip}
\end{table}

Figure~\ref{fig:SEAvsSDEAspring} presents a comparison of the linear spring rendering performance of SEA and SDEA when the controller gains are selected to be the same. Figure~\ref{fig:SEAvsSDEAspring} indicates that the performance of SEA and SDEA are close to each other in the low frequency range, while the transition from spring rendering to high frequency dynamics differs significantly. % between SEA and SDEA.
%As can be verified from Table~\ref{table:numericSEASDEA}, the inerter effects are lower for SEA, compared to SDEA, while the low frequency damping effects are identical. %Table~\ref{table:massdamperadd} presents the numerical values for the elements that provide the frequency dependent damping effect. 

For Voigt model rendering,  the realizations in Table~\ref{fig:allresults}b and~\ref{fig:allresults}e  for SEA and SDEA under VSIC have distinct topologies, as they are valid for non-overlapping parameter ranges. The $K_{vir}$-$B_{vir}$ plot for SEA is depicted in Figure~\ref{fig:KBplot}, where the boundaries of the plot are $\frac{\alpha}{\alpha +1}\,K_{ref}$ and $\frac{\alpha}{\alpha +1}\,B_{ref}$, respectively. Figure~\ref{fig:KBplot} indicates that the selection of higher $B_{ref}$ values allows for passive rendering of lower $K_{vir}$ levels. 

The $K_{vir}$-$B_{vir}$ plot of SDEA under VSIC during Voigt model rendering is presented in the Figure~\ref{fig:KBplotSDEA}, for the case when the system parameters are such that Condition~(iv) of Proposition~\ref{theorem:SDEAPPvoigttheorem} is more conservative, as is the case for the experimental setup in Section~\ref{sec:ExperimentalVal}. Alternatively, if the system parameters are such that Condition~(i) of Proposition~\ref{theorem:SDEAPPvoigttheorem} is the more conservative, then  $K_{vir}$ increases with higher $B_{vir}$ as presented in the Supplementary Document~\cite{supplementary}.   

Due to the page restrictions, numerical analyses of the effects of the plant parameters and the controller gains on the rendering performance are presented in the Supplementary Document~\cite{supplementary} through a comprehensive set of Bode plots.

%Null impedance rendering performance comparison of SEA and SDEA is presented in the Supplementary Document~\cite{supplementary}.  

%In this section, we study the effects of the physical plant parameters and the controller gains on the rendering performance through Bode plots. %We  provide performance comparisons between SEA and SDEA  using the insight gained through the passive realizations of the closed-loop systems. 
 %We also numerically evaluate the effective impedance of closed-loop systems.  Only sample Bode plots in an order of increasing complexity are presented within the manuscript, for the brevity of discussion. The complete set of results is available in the Supplementary Document~\cite{supplementary}. %and these evaluations have been checked by bode plot of output impedance transfer functions. Performance of the systems which have been analysed in Section~\ref{sec:SEApassivity} and Section~\ref{sec:SDEApassivity} are investigated.

  %The passive physical equivalents in Table~\ref{fig:allresults} explicitly show that when causal controllers roll-off, the dynamics of the uncontrolled plant is recovered. Accordingly, the high frequency interaction port impedances of all realizations are dominated by the  dynamics of the  physical filters (spring and damper) serially attached to the plant; hence, passive physical realizations indicate that a SEA acts as a physical spring $K$, while an SDEA acts as a physical spring $K$ and damping $B_f$ in parallel, at high frequencies. 

% At high frequencies, all causal controllers roll off and independent of the controller, 

\vspace{-3mm}
\section{Co-Design via Passive Physical Equivalents} % and Passivity Bounds for SEA and SDEA}

The physical plant parameters are crucial as they determine the limits of haptic rendering performance under passivity constraints~\cite{newman_1992,Colgate_realizations,kenanoglu2023}. When the causal controllers roll-off, the dynamics of the uncontrolled plant are recovered for all closed-loop systems. %Hence, the high-frequency interaction port impedance of all S(D)EA realizations under causal controllers is dominated by the  dynamics of the  physical filters serially attached to the plant. 
 Accordingly, when the controller gains are set to zero in the realizations, SEA acts as a physical spring $K$, while an SDEA acts as a physical spring-damper $K$-$B_f$ pair at high frequencies as seen from the interaction port.

Given passive physical equivalents do not distinguish between the plant parameters and the controller gains, they promote co-design thinking by enforcing simultaneous and unbiased consideration of controller and plant dynamics on the closed-loop system performance~\cite{Colgate_realizations,Kamadan2017,Kamadan2019}.  For instance, in terms of rendering fidelity, the passive physical equivalents of SEA and SDEA under VSIC while rendering Voigt, linear spring, and null impedance models indicate that the selection of higher controller gains has the same effects as employing a plant with lower inertia and damping, as the controllers can compensate for the plant dynamics up to their control bandwidth. However, lower inertia and damping parameters of the plant necessitate lower controller gains to ensure passivity.  

Since the plant damping is commonly considered as a parasitic effect, passive Voigt model rendering with SEA has gone unnoticed in the literature, until this study, where passive realizations are considered for the analysis. A close inspection of the  passive physical equivalent of SEA during Voigt model rendering presented in Table~\ref{fig:allresults}e indicates that higher virtual damping $B_{vir}$ levels can be  passively rendered if %$c_{2v}~=~\frac{B_m +G_m +B_{ref} \alpha }{\alpha +1}-\frac{\alpha \,K_{ref} \,{\left(B_m +G_m \right)}}{K\,{{\left(\alpha +1\right)}}^2 }$ 
 $c_{2v}$ can be set high. The upper bound of $c_{2v}$ is imposed by $\frac{B_m+G_m}{(\alpha+1)}$, as $B_{ref}$ needs to be negative for the passivity. Hence, if the upper bound on passive damping rendering is to be increased, then one can employ a plant with higher $B_m$.
 
This motivates the intentional addition of electrical damping to the system to augment the motor damping; a method commonly employed for sampled-data passivity of impedance-type haptic interfaces~\cite{Weir2008}. Utilizing a plat with higher $B_m$ not only enables the passive rendering of higher virtual damping but also relaxes the bounds on the virtual stiffness, enlarging the  $K_{vir}$-$B_{vir}$ plot, as depicted in Figure~\ref{fig:KBplot}. Furthermore, since the rendered damping becomes more coupled to the interaction port as the stiffness of the filter gets higher, one can increase $K$ to improve the damping rendering performance. %~of~SEA.

\vspace{-3mm}
\section{Experimental Validation} %\vspace{-1mm}
\label{sec:ExperimentalVal} %
In this section, we experimentally validate the theoretical passivity bounds and the haptic rendering performance of S(D)EA using a customized version of the SEA brake pedal presented in~\cite{caliskan2018series,caliskan2020efficacy}.  %, is actuated by a brushless DC motor equipped with a Hall-effect sensor and an optical encoder.  The torque output of the motor is amplified with a 1:39.5 transmission ratio. The series elastic element is implemented as a linear spring through a compliant cross-flexure joint embedded into the capstan pulley. The  deflections of the cross-flexure joint are measured with a linear encoder to estimate the interaction torques. All controllers are implemented in real-time at 1~kHz utilizing an industrial PC connected to an EtherCAT bus.
To implement a series \emph{damped} elastic brake pedal, linear eddy current damping is added in parallel to the compliant element of the SEA brake pedal. In particular, permanent magnets arranged as a Halbach array to augment the magnetic field are placed to face an aluminum plate. The distance between the magnet array and the aluminum plate is adjusted to control the viscous damping added to the system. When the magnets are removed, the SDEA pedal simplifies to an SEA.  A video of the SDEA pedal is available as the Multimedia Extension. %~\cite{jang2002characteristic,yi2010research} 
%  distance is 12 mm.
  
%encoder resolution  is 1024 count/rev on the shaft
\begin{comment}
\begin{figure}[h]
  \centering  %\vspace{-4mm}
  \includegraphics[width=0.45\columnwidth]{figures/brakepedal3_v1.jpg}
  \vspace{-2mm}
    \caption{The S(D)EA brake pedal} %\vspace{-5.5mm}
    \label{fig:breakpedal}  \vspace{-0.9\baselineskip}
\end{figure}
\end{comment}

\begin{comment}
    \begin{figure*}[b]
\centering
\includegraphics[width=2.05\columnwidth]{figures/Passivity_bounds_verification.pdf} \vspace{-2mm}
\vspace{-.95\baselineskip}
\caption{Passivity bounds vs experimentally determined coupled stability: \sout{(a) $I_m$--$G_t$ plot for SEA during null impedance rendering,} (b) $G_t$--$K_{vir}$ plot for SEA  during spring rendering, and (c) $G_t$--$K_{vir}$ plot for SDEA during spring rendering.}
\label{fig:passivityresults}
%\vspace{-1\baselineskip}
\end{figure*}
\end{comment}

 \vspace{-3mm}
\subsection{Identification of Plant Parameters}

%An accurate determination of the plant parameters is important to verify the passivity bounds of the system. %First, the stiffness and damping of the physical filter are identified. %, respectively. 
The stiffness of the cross-flexure joint and the Eddy-current damping due to the magnet array are experimentally determined as $K$ = 121.8~N-m/rad and $B_f$ = 0.0127 N-m s/rad, respectively. 
%
%by applying pre-determined torques  to the end-effector and measuring the resulting deflections when the actuator is locked. A least square fit to the experimental data indicates that $K$ = 121.8~N-m/rad with $R^2$~=~0.99. 
%
%For the identification of the damping, the magnet array is fixed to a force sensor and the motion of the aluminum plate is %controlled to follow a reference chirp signal with an amplitude of 33.5~rad/sec over the range of 0.9-1.4~Hz. % The velocity is estimated through the numerical differentiation of the encoder data. % using the curve braking velocity estimator~\cite{ghaffari2013high}.
%A least square fit indicates that $B_f$ = 0.0127 N-m s/rad with $R^2$~=~0.84. 
%
 Closed-loop system identification is utilized to experimentally determine the system parameters related to the motor and the power transmission. The closed-loop identification enables accurate prediction of the plant parameters using LTI techniques since the robust motion controller effectively compensates for the hard-to-model nonlinear effects in the power transmission. To determine the reflected inertia and damping of the plant, the system identification is performed under the inner velocity controller with  $G_m$~=~0.0576~N-m~s/rad.  %A chirp velocity reference signal with an amplitude of 7.85~rad/sec is applied to the motion control loop over the  range of 0.001-10~Hz, while no exogenous torque $\tau_{sea}$ is applied to the system. 
  A first-order transfer function is fitted to the data to determine the plant parameters as $J_m$~=~0.0024~kg-m$^2$ and $B_m$~=~0.0177~N-m~s/rad with $R^2$~=~0.88. 

For simplicity of presentation, the theoretical passivity bounds have been derived under the non-limiting assumption that the power transmission of the system has a unity reduction ratio. %On the other hand, the S(D)EA brake pedal has a reduction ratio $n$ of 39.5. 
 Equivalent plant parameters and controller gains can be established for systems with a reduction ratio of $n = 39.5$ by introducing the following mappings: $J_{m_{eq}} = n^2 \: J_m $, $B_{m_{eq}} = n^2 \: B_m$, and $G_{m_{eq}} = n^2 \: G_m $, and $G_{t_{eq}}= 1/n \: G_t$. Unless otherwise stated, the controller gains of~VSIC are set to $G_m$~=~0.0576~N-m~s/rad and $G_t$~=~25~rad/(s~N-m), respectively.

%\begin{figure}[h]
%  \centering 
%  \includegraphics[width=0.7\columnwidth]{figures/SystemId.pdf} \vspace{-2mm}
%     \captionsetup{justification=centering}
%    \caption{Closed loop motion controlled system}
%    \label{fig:SystemId} \vspace{-2mm}
%\end{figure}

%\begin{equation}
%    \frac{\omega_m}{\omega_d}=\frac{G_m}{J_m s + B_m + G_m}
%    \label{eqn:sistemid}
%\end{equation}
%We know the $G_m$ in this model, and the rest of physical parameters can be found with the estimated transfer function.

%Table~\ref{table:systemId1} presents the plant parameters of the S(D)EA brake pedal. 

\begin{figure*}[t]
\centering %\vspace{-.3\baselineskip}
\begin{tabular}{cc}
  \begin{subfigure}[a]{0.285\textwidth}
   \includegraphics[keepaspectratio=true, width=\linewidth]{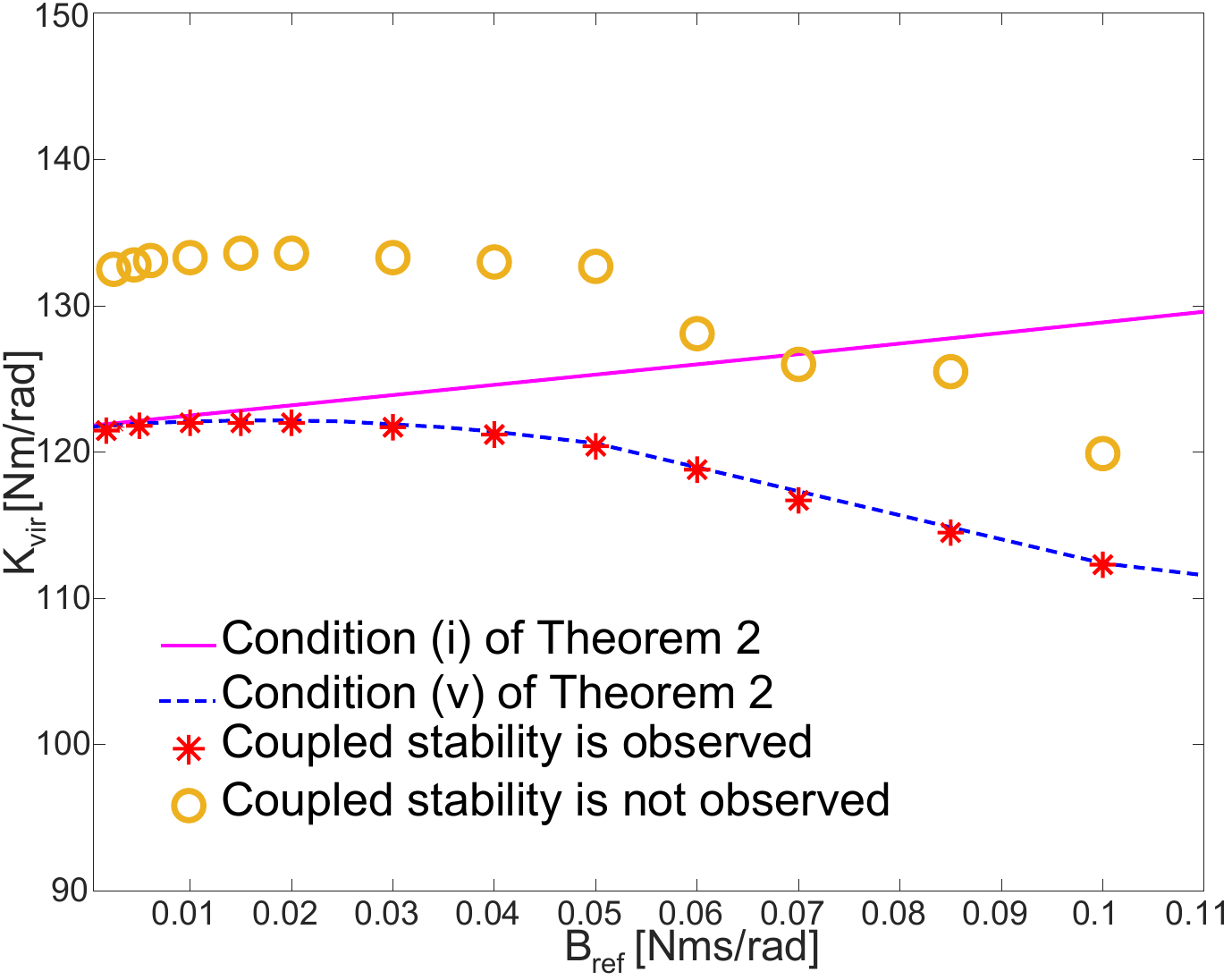}
     \vspace{-6mm}
    \caption{SDEA -- Voigt rendering}
    \label{fig:SDEApassivityVoigt}
    \end{subfigure} &
    \begin{subfigure}[a]{0.285\textwidth}
   \includegraphics[keepaspectratio=true, width=\linewidth]{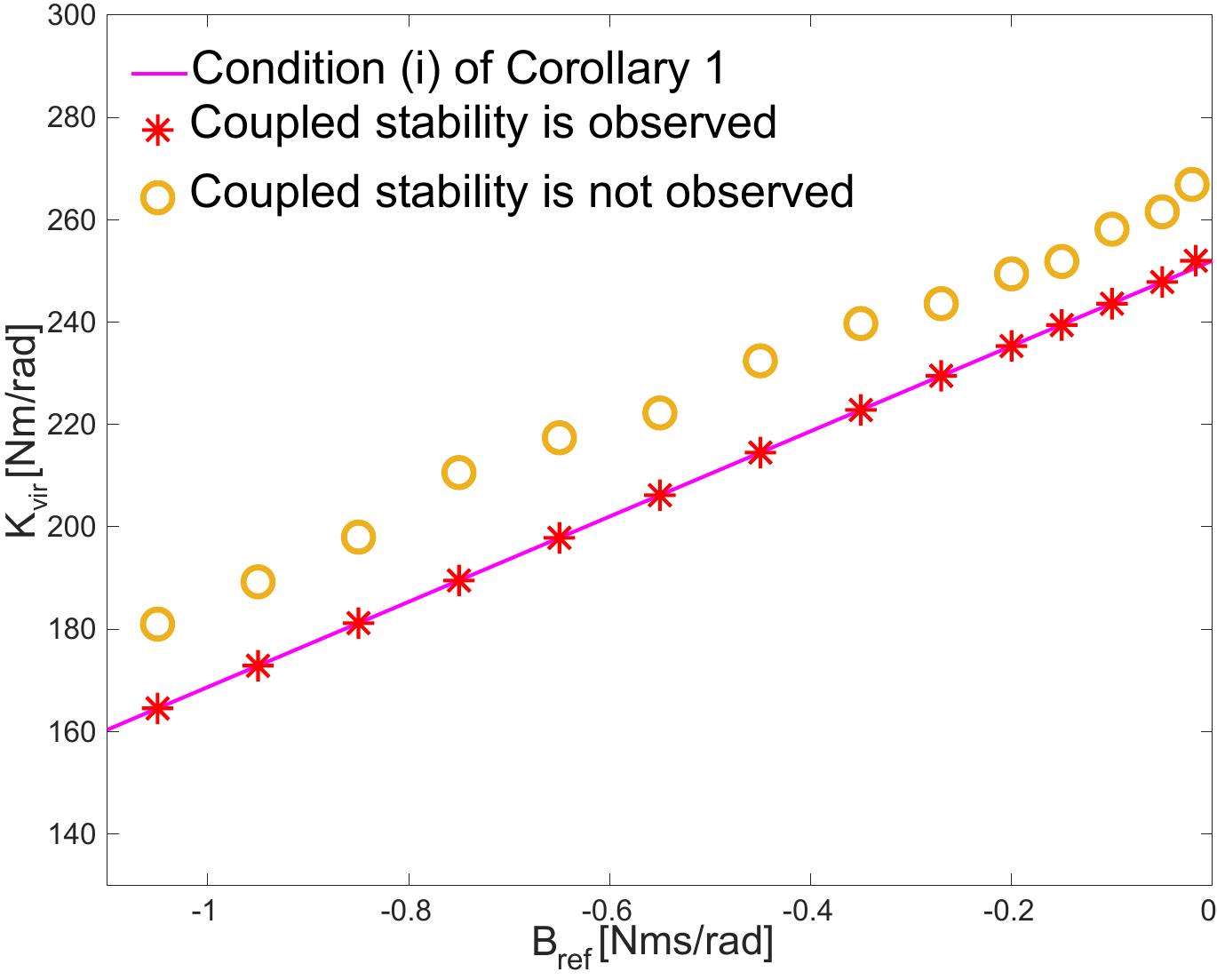}
     \vspace{-6mm}
    \caption{SEA -- Voigt  rendering}
    \label{fig:SEApassivityVoigt}
    \end{subfigure} 
\end{tabular}    
\begin{tabular}{cc}    
      \begin{subfigure}[a]{0.285\textwidth}
  \includegraphics[keepaspectratio=true, width=\linewidth]{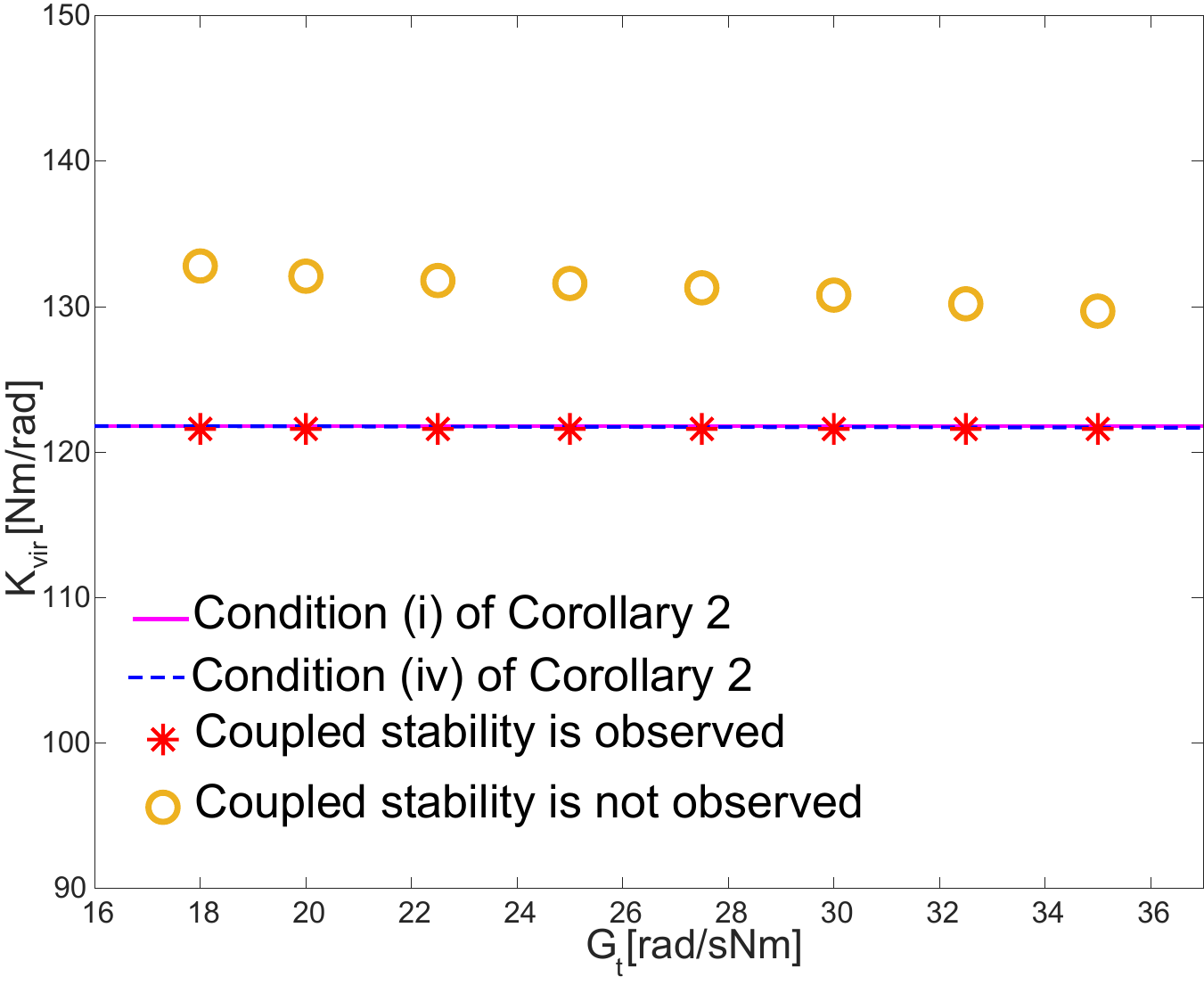}
              \vspace{-6mm}
    \caption{SDEA -- spring rendering}
    \label{fig:SDEApassivityspring}
    \end{subfigure} &
 \begin{subfigure}[a]{0.285\textwidth}
  \includegraphics[keepaspectratio=true, width=\linewidth]{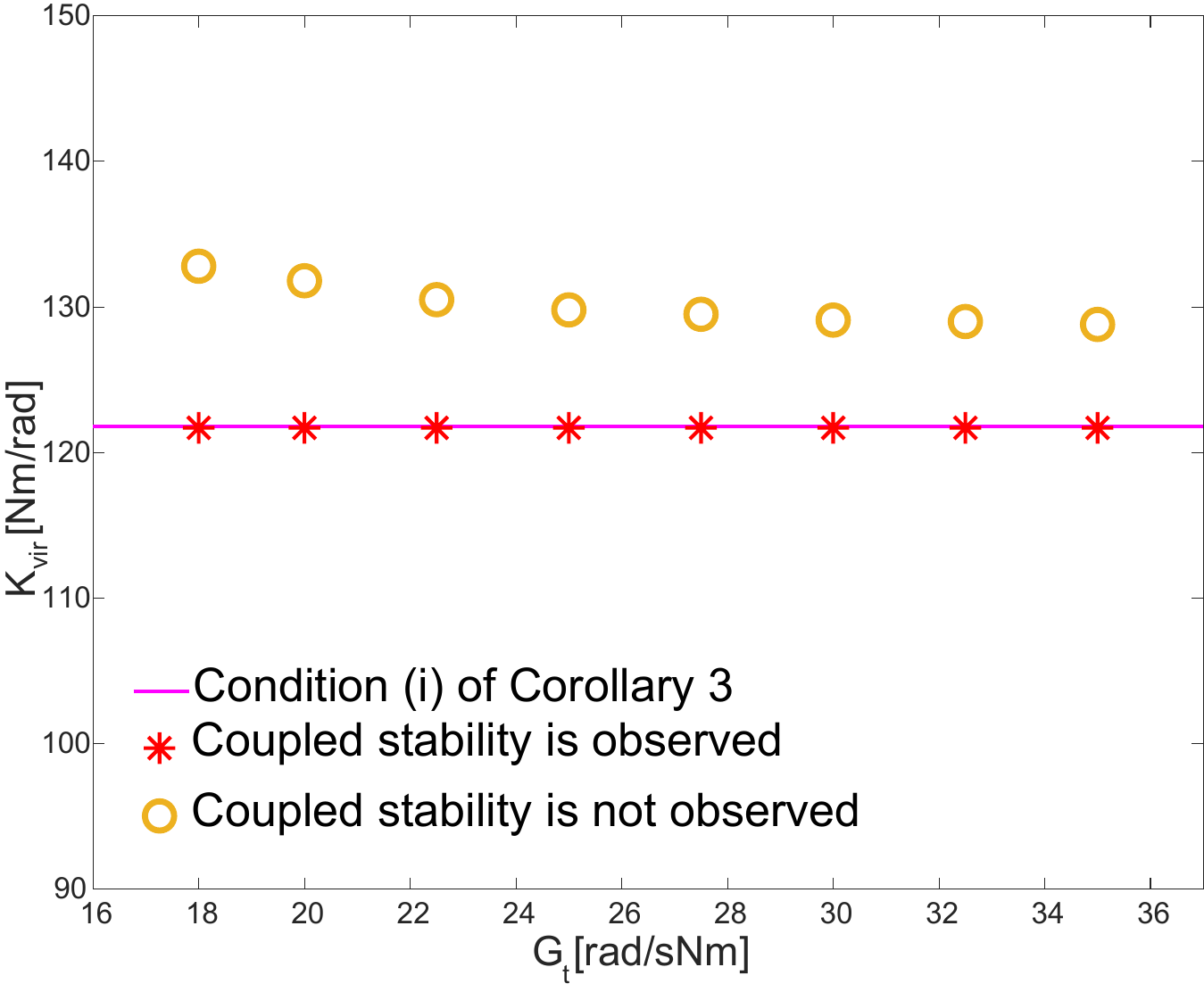}
               \vspace{-7mm}
    \caption{SEA -- spring rendering}
    \label{fig:SEApassivityspring}
    \end{subfigure}  
\end{tabular}      
    \vspace{-2mm}
    \caption{Passivity bounds vs experimental coupled stability for SDEA/SEA during Voigt model and spring rendering}
        \label{fig:passivityresults}
        \vspace{-1.25\baselineskip}
  \end{figure*}
  
 \vspace{-4mm}
\subsection{Verification of Passivity Bounds} % \vspace{-.85mm}

%The challenging nature of the experimental determination of closed-loop system phase with high accuracy and repeatability renders the empirical verification of system passivity through Bode phase plots unreliable. On the other hand, 

It has been established in the literature that the passivity of a system can be investigated by studying the coupled stability of interactions when the system is exposed to the most destabilizing environments~\cite{colgate_1988}. In particular, passivity can be concluded if and only if there exists no set of ideal springs or inertias that destabilize the system under excitations that span the whole frequency spectrum~\cite{Tosun2020}. For  SEA, inertial environments have been determined to be among the most destabilizing~\cite{Losey2017}.

To validate the theoretical passivity bounds established in this paper, the S(D)EA brake pedal is coupled to a range of inertias, and impacts are imposed to the end-effector to excite the system at all possible frequencies. If the violation of coupled stability (e.g., chatter) is observed, then the system is not passive. If no violation of the coupled stability is observed after many trials during which the end-effector inertia of the S(D)EA brake pedal is gradually increased, then it is concluded that the experimental evidence indicates the passivity of the system. A video of the coupled stability experiments is provided in the Multimedia Extension.

%\begin{table}[t] 
%\caption{Parameters of the S(D)EA brake pedal} \vspace{-2mm}
%\centering 
%\begin{tabular}{|c| c|}
%\hline
%Plant Parameter & Value\\ 
%\hline 
%$J_m$ & $0.0024 $ kg-m$^2$ \\  [0.1ex]
%\hline
%$B_m$ & $0.0177$ N-m s/rad \\ [0.1ex]
%\hline
%$K$   & $121.8$ N-m/rad \\ [0.1ex]
%\hline
%$B_f$ & $0.0127$ N-m s/rad \\ [0.1ex]
%\hline
%\end{tabular} \vspace{-5mm}
%\label{table:systemId1} 
%\end{table}

\smallskip

\noindent \underline{Voigt Model Rendering with SDEA:} In this experiment, we have investigated the coupled stability of SDEA under VSIC during Voigt model rendering when the controllers are~P. To validate with the necessary and sufficient conditions provided in Proposition~\ref{theorem:SDEAPPvoigttheorem}, we have tested various $K_{ref}$ and $B_{ref}$ values. 

Figure~\ref{fig:SDEApassivityVoigt} depicts the $K_{vir}$--$B_{ref}$ plot obtained from the experiments conducted on the brake pedal with SDEA under VSIC during Voigt rendering, where $K_{vir} = \frac{\alpha}{\alpha+1} K_{ref}$. The magenta and blue lines in the figure represent the theoretical passivity bound according to Conditions (i) and~(v) of Proposition~\ref{theorem:SDEAPPvoigttheorem}, respectively. The symbols ``*" and ``o" indicate the experiments where coupled stability was preserved and compromised, respectively. The experimental results confirm the analytically predicted passivity boundaries. The experimental values are in good agreement with the theoretical values, with an error of approximately 8\%. The experimental results may be slightly more conservative due to unmodelled friction and hysteresis effects, which cause extra dissipation. 

\smallskip

\noindent \underline{Voigt Model Rendering with SEA: } In this experiment, we have investigated the coupled stability of SEA under VSIC during Voigt model rendering when the controllers are~P. To validate with the necessary and sufficient conditions provided in Corollary~\ref{theorem:SEAPPvoigttheorem}, we have tested various $K_{ref}$ and $B_{ref}$ values when $G_t$~=~15~rad/(s~N-m). %Please note that the stiffness of the physical spring is equal to 252 Nm/rad for all Voigt model rendering experiments with SEA.

Figure~\ref{fig:SEApassivityVoigt} depicts the $K_{vir}$--$B_{ref}$ plot obtained from the experiments conducted on the brake pedal with SEA under VSIC during Voigt model rendering. In the figure, the theoretical passivity bound according to Condition (i) of Corollary~\ref{theorem:SEAPPvoigttheorem} is depicted as the magenta line. The experimental results validate the analytically predicted passivity boundary and the theoretical bound is determined to be 7\% more conservative. %, as the physical system is likely to have some extra dissipation  due to unmodelled dissipation effects.
\smallskip

\begin{figure*}[t]
\centering   \vspace{-.5\baselineskip}
    \includegraphics[width=1.9\columnwidth]{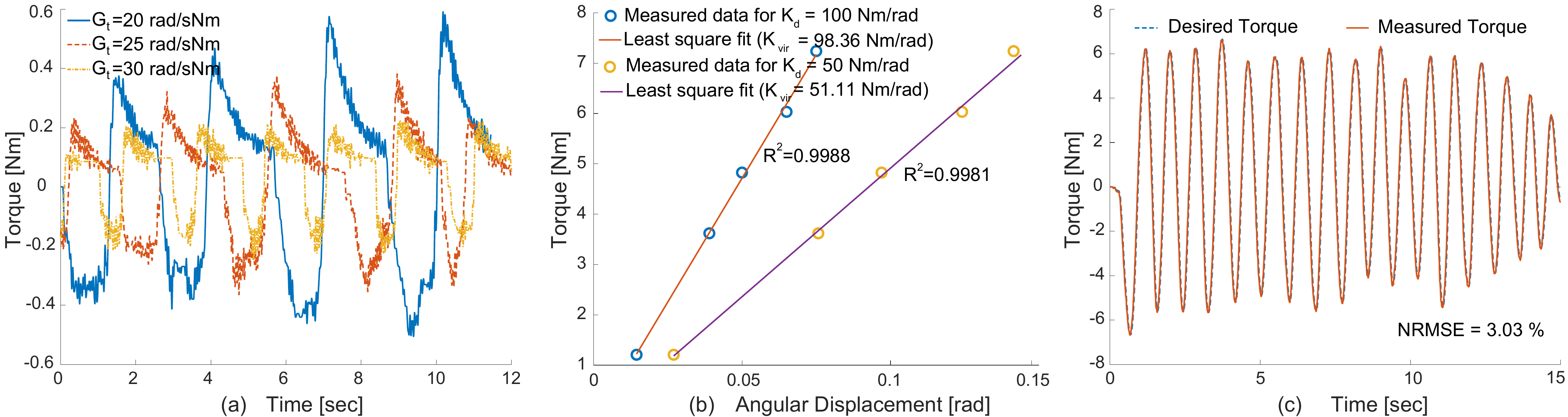} 
    \vspace{-.6\baselineskip}
    \caption{(a) Null impedance rendering performance (potato chip) test for $G_t$~= 20,~25 and~30~rad/(s~N-m), (b) Virtual stiffness rendering for $K_{ref}$~=~50 and~100~N-m/rad when $G_t$~=30~rad/(s~N-m), and (c) Tracking performance of SEA during virtual spring rendering with $K_{ref}$~=~100~N-m/rad and $G_t$~=~30~rad/(s~N-m).}
    \label{fig:renderingperformance}
    \vspace{-1.1\baselineskip}
\end{figure*}

\noindent \underline{Spring Rendering with SDEA:}
In these experiments, we have studied the coupled stability of SDEA under VSIC during spring rendering when both controllers are P. We have selected one passive and one active $K_{ref}$ values for eight distinct $G_t$ gains according to theconditions given in Conditions (i) and~(iv) of Proposition~\ref{theorem:SDEAPPvoigttheorem} when $B_{ref}=0$. 

Figure~\ref{fig:SDEApassivityspring} presents the experimental $K_{vir}$--$G_t$ plot for the SDEA brake pedal. In the figure, the theoretical passivity bound according to Condition (i) of Proposition~\ref{theorem:SDEAPPvoigttheorem} when $B_{ref}=0$ is depicted as the magenta line and equal to physical stiffness of the SDEA, while  the bound according to Condition (iv) of Proposition~\ref{theorem:SDEAPPvoigttheorem} when $B_{ref}=0$ is depicted as the blue line. Figure~\ref{fig:SDEApassivityspring} shows that the two conditions are very close to each other for the parameters of the SDEA brake pedal. The experimental results validate the analytically predicted passivity boundary and the theoretical bound is determined to be about 6.5\% more conservative. %, as the physical system is likely to have some extra dissipation due to unmodelled dissipation effects.

%Figure \ref{fig:springexprimentpassivitySDEA} shows that under the condition lines points hold coupled stability, but $K_{vir}$ is over the condition lines, the chatter was observed at end-effector, so the coupled stability is violated. Figure \ref{fig:springexprimentpassivitySDEA} presents there is a difference between the condition lines and the points where coupled stability is violated. It can be caused by the omitted damping on the end-effector.

\smallskip 

\noindent \underline{Spring Rendering with SEA:}
In these experiments, we have studied the coupled stability of SEA under VSIC during spring rendering when both controllers are P. We have selected one passive and one active $K_{ref}$ values for eight distinct $G_t$ gains according to the necessary and sufficient condition given in Condition (i) of Corollary~\ref{theorem:SEAPPvoigttheorem} when $B_{ref}=0$.

Figure~\ref{fig:SEApassivityspring} presents the experimental $K_{vir}$--$G_t$ plot for the SEA brake pedal. In the figure, the theoretical passivity boundary is depicted as the magenta line and equal to physical stiffness of the SEA  according to Condition (i) of Corollary~\ref{theorem:SEAPPvoigttheorem} when $B_{ref}=0$. The experimental results validate the analytically predicted passivity boundary and the theoretical bound is determined to be about 7\% more conservative. %, as the physical system is likely to have some extra dissipation  due to unmodelled dissipation effects.

\smallskip

\vspace{-5mm}
\subsection{Evaluation of Haptic Rendering Fidelity}

In this subsection, we have experimentally evaluated the performance of S(D)EA under VSIC during rendering Voigt, spring, and null impedance models. Since the haptic rendering performance of SDEA under VSIC is very similar to that of SEA for the experimental setup, only the results for SEA are provided for the brevity of the presentation.

%For indicating a parasitic torque $1.48\%$ for $G_t=20$ rad/sNm, $0.95\%$ for $G_t=25$ rad/sNm, and $0.62\%$ for $G_t=30$ rad/sNm.       

%One of the good experiments of null impedance rendering is chip test. In this test, human pushed the brake pedal with a chip without breaking it. Figure~\ref{fig:chiptestofSEA} presents results of the chip test.

%Figure \ref{fig:chiptestofSEA} shows that brake pedal can be pushed by chip without breaking it between %0.24--0.59 Nm which is the acceptable results as the maximum output torque capability of the is nearly 40 Nm. In sections~\ref{sec:SEApassivity} and~\ref{sec:Rendering Performance}, we have shown that higher $G_t$ increases null impedance rendering performance. Figure~\ref{fig:chiptestofSEA} shows that Therefore, this experimental results verify the result of the sections~\ref{sec:SEApassivity} and~\ref{sec:Rendering Performance}. Higher force/torque controller gain results in better null impedance rendering performance. 

\smallskip
\noindent \underline{Null Impedance Rendering with SEA:}
The performance of SEA under VSIC during null impedance rendering is important, as this control mode provides active backdrivability to allow users to move the system without much resistance.

Figure~\ref{fig:renderingperformance}a presents the null impedance rendering performance of SEA under VSIC for three distinct levels of the torque controller gain $G_t$. As the torque controller gain $G_t$ is increased from 20~rad/(s~N-m) to  30~rad/(s~N-m), the torque required to move the pedal decreases from 1.48\% to 0.62\% of 40~N-m torque output capacity of the SEA brake pedal. %0.59~N-m to  0.24~N-m
 Note that this level of active-backdrivability is excellent for the SEA brake pedal, as evidenced by a commonly employed chip test (please refer to the Multi-Media Extension), where a potato chip is used to move the device  without getting broken. The experimental results in Figure~\ref{fig:renderingperformance}a are also in good agreement with the analysis presented in \cite{supplementary}, where the positive effect of increasing the torque controller gain $G_t$ on the null impedance rendering performance has been shown. 
 
\smallskip

\noindent \underline{Spring Rendering with SEA:}
The performance of SEA under VSIC  during spring rendering is important, as this control mode is commonly used to implement virtual constraints to avoid users to reach the undesired regions of the workspace.
Figure~\ref{fig:renderingperformance}b presents the experimental verification of the spring rendering performance for two distinct levels of virtual stiffness, where $K_{ref}$~=~50 and~100~N-m/rad when $G_t$~=~30~rad/(s~N-m). The rendered stiffness of the SEA under VSIC  is experimentally determined by applying pre-determined torques to the end-effector and measuring the resulting deflections. A least square fit to the experimental data indicates that $K_{vir}$~=~51.11~N-m/rad with $R^2$~=~0.99 and $K_{vir}$~=~98.36~N-m/rad with $R^2$~=~0.99, resulting in 3.73\%  error for $K_{ref}$~=~50~N-m/rad and 0.19\% error for $K_{ref}$~=~100~N-m/rad, respectively. These experiments are repeated for $G_t$~=~25~rad/(s~N-m). In this case, a least square fit to the experimental data indicates that $K_{vir}$~=~51.27~N-m/rad with $R^2$~=~0.99 and $K_{vir}$~=~98.99~N-m/rad with $R^2$~=~0.99, resulting in 4.35\%  error for $K_{ref}$~=~50~N-m/rad and 0.73\% error for $K_{ref}$~=~100~N-m/rad, respectively. The experimental results in Figure~\ref{fig:renderingperformance}b are in good agreement with the analysis presented in \cite{supplementary}, where the positive effects of increasing the torque controller gain $G_t$ and desired virtual stiffness $K_{ref}$ on the spring rendering performance have been shown. 

Figure~\ref{fig:renderingperformance}c presents the interaction performance of the SEA brake pedal under dynamic inputs from a user.  During these experiments, the SEA brake pedal was rendering a virtual stiffness of $K_{ref}$~=~50~N-m/rad with $G_t$~=~30~rad/(s~N-m). The desired interaction torque due to the virtual stiffness model and the interaction torques estimated through the series elastic element are presented with dashed and solid lines, respectively. The normalized RMS error for this dynamic tracking task is computed as 3.03\%. This experiment was also repeated for $G_t$~=~25~rad/(s~N-m). In this case, the normalized RMS error is computed as 3.28\%. These experimental results are in good agreement with the analysis presented in \cite{supplementary}, where it has been demonstrated that increasing the torque controller gain $G_t$ improves the force tracking performance.

\smallskip

\noindent \underline{Voigt Model Rendering with S(D)EA:}
The interaction performance of the brake pedal under dynamic inputs from a user is studied for SDEA during Voigt model rendering. The interaction performance is experimentally verified while SDEA brake pedal is rendering a Voigt model with $K_{ref}=$ 100~Nm/rad and $B_{ref}=$  0.01~Nms/rad. The desired interaction torque due to the virtual Voigt model is computed and compared with the interaction torques estimated through the series damped elastic element. These experimental results have a normalized RMSE of 1.3\%, validating the fidelity of Voigt model rendering. % task with SDEA.

A similar experiment has also been conducted for the experimental verification of the interaction performance of the SEA brake pedal under dynamic inputs from a user while rendering a Voigt model with  $K_{ref} \!=\! $ 100~Nm/rad and $B_{vir}\!=\!$ 0.01~Nms/rad. The normalized RMSE  is computed as 4.3\%, indicating high fidelity for the Voigt model rendering task. % with~SEA. 

%Figure~\ref{fig:expperformance}c depicts the desired interaction torques derived from the Voigt model and the corresponding interaction torques measured using the series damped elastic element, denoted by dashed and solid lines, respectively. 

\vspace{-3mm}
\section{Conclusions and Discussion} \vspace{-1mm}
\label{sec:discusion}

We have derived minimal passive mechanical equivalents for these systems to provide intuition into their closed-loop dynamics. The passive mechanical equivalents make the control authority and parasitic dynamics of the system explicit and enable the rigorous study of system parameters and controller gains on the rendering performance. The passive mechanical equivalents provide a concrete understanding of the limitations of rendering performance (e.g., the stiffness of the physical stiffness provides an upper bound on virtual spring rendering under VSIC). These results significantly extend the interaction control analyses in~\cite{Colgate_realizations,kenanoglu2023}  to S(D)EA and provide insights into the robust stability-transparency trade-off.

%passivity bounds established and experimentally verified in~\cite{Tosun2020,Mengilli2020,mengilli2021passivity}. %vallery2007passive,vallery2008compliant,tagliamonte2014passivity,calanca2017impedance,
% Mengilli2021 SDEA oneport WHC paper

We have also demonstrated that passive mechanical equivalents enable fair comparisons of different plants (e.g.,  SEA vs SDEA) on the haptic rendering performance. Unlike the case in numerical studies, comparisons of closed-loop system dynamics through passive physical equivalents are informative in that these conclusions can be generalized. These comparisons highlight the impact of different plant and controller terms on the closed-loop rendering performance. Furthermore, since there exists continuity among realizations, the effect of each controller term on plant dynamics can be rigorously studied. Moreover, these comparisons are symbolic in nature and do not require performance optimization of each closed-loop system to ensure fairness, as emphasized in~\cite{Yusuf_multi_2020}.

We have also emphasized that passive mechanical equivalents provide a physical realization of the effective impedance, establishing an intuitive understanding of the effective impedance analysis. For instance, realizations show how a frequency-dependent damping effect in the effective impedance analysis can be realized with a serial connection of an inerter with a damper, as in~\cite{mehling2005increasing}. 

%In Section~\ref{sec:Rendering Performance}, we have presented comprehensive simulations to further demonstrate the effect of controller and system parameters on the haptic rendering performance of S(D)EA under VSIC. 
We have advocated that passive physical equivalents promote co-design by enabling concurrent consideration of plant parameters and controller gains on the haptic rendering performance. The realization of Voigt model rendering with SEA is provided as an illustrative example that demonstrates how the plant damping can be augmented and negative controller gains can be employed to achieve a larger range of passively renderable virtual environments.

In addition to the passivity physical equivalents, we have also presented the passivity analysis of SEA and SDEA under VSIC while rendering Voigt models, linear springs, and the null impedance, and provided the necessary and sufficient conditions for the passivity of these systems. Our results significantly extend the results on S(D)EA passivity in the literature~\cite{Tosun2020,Mengilli2020,mengilli2021passivity}, by providing the necessity of the conditions and allowing the controller gains to be negative, and enabling passive Voigt model rendering with SEA under~VSIC. 

%We have presented passive physical realizations for S(D)EA under VSIC \textcolor{blue}{while rendering Voigt models,  ideal springs, and the null impedance} in Table~\ref{fig:allresults}.

It is important to note that, in general, passive physical realizations for a given impedance transfer function are not unique. While the feasibility conditions for a passive physical realization provide sufficient conditions for passivity as shown in Section~\ref{sec:SEApassivity}, the necessity cannot be easily established through such analysis, as it requires studying the feasibility of \emph{all} passive physical realizations.

%\begin{figure}[b]
%\centering \vspace{-2mm}
 %     \begin{subfigure}[a]{0.35\linewidth}
 % \includegraphics[keepaspectratio=true, width=\linewidth]{figures/SEA_null_PI_P_spring_v2.pdf}
  %            \vspace{-5mm}
  %  \caption{ }
  %  \label{fig:SEAnullPIP2}
   % \end{subfigure} 
%          \begin{subfigure}[a]{0.34\linewidth}
%  \includegraphics[keepaspectratio=true, width=\linewidth]{figures/SEA_null_PI_P_lever.pdf}
 %             \vspace{-5mm}
  %  \caption{ }
   % \label{fig:SEAnullPIP3}
%    \end{subfigure} 
 %           \vspace{-2mm}
 %   \caption{Alternative passive physical equivalents of SEA under VSIC during null impedance rendering when the force controller is P and the motion controller is~PI}
  %  \label{fig:SEAnullPIPAlternative}
       % \vspace{-6mm}
  %\end{figure}
  
  %  \begin{figure}[t]
%\centering \vspace{-1mm}
 %     \begin{subfigure}[a]{0.35\linewidth}
 % \includegraphics[keepaspectratio=true, width=\linewidth]{figures/SDEA_Spring_P_P_spring.pdf}
  %            \vspace{-5mm}
   % \caption{ }
%    \label{fig:SEAnullPIP2}
 %   \end{subfigure} 
  %        \begin{subfigure}[a]{0.35\linewidth}
  %\includegraphics[keepaspectratio=true, width=\linewidth]{figures/SDEA_Spring_P_P_lever.pdf}
   %           \vspace{-5mm}
   % \caption{ }
%    \label{fig:SEAnullPIP3}
 %   \end{subfigure} 
 %           \vspace{-3mm}
 %   \caption{Alternative passive physical equivalents of SDEA under VSIC during spring  rendering when both controllers are~P}
  %  \label{fig:SDEAspringPPAlternative}
  %      \vspace{-6mm}
  %\end{figure}

\begin{figure}[t]
  \centering % \vspace{-4mm}
  \includegraphics[width=\linewidth]{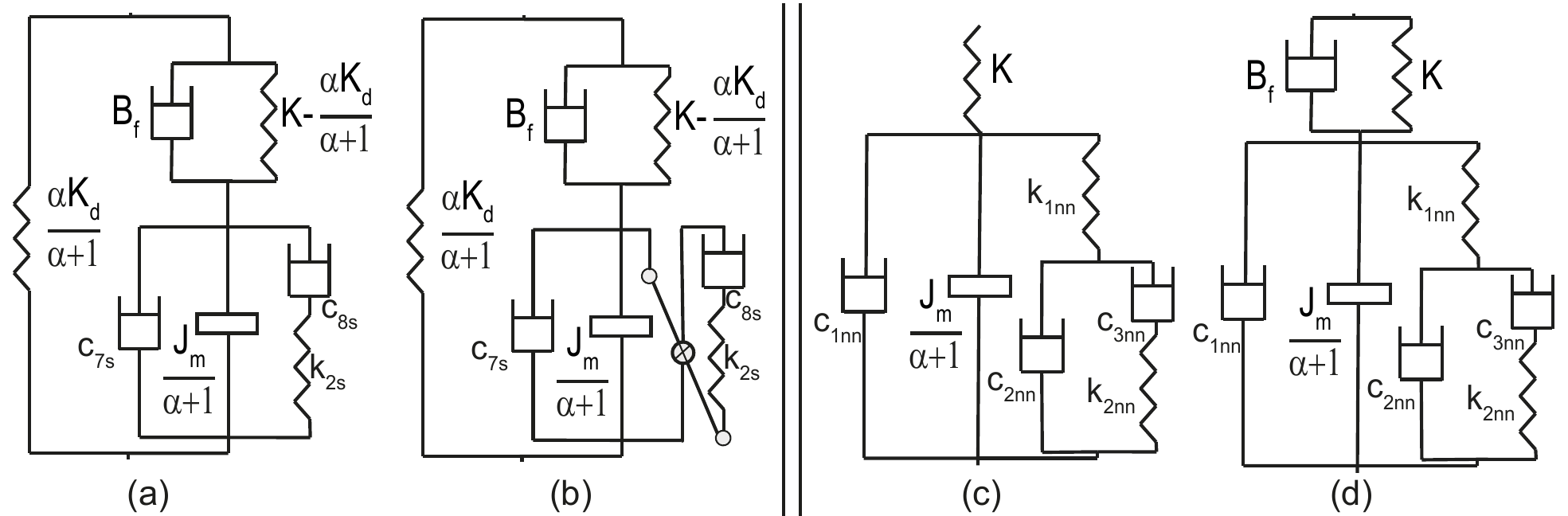}
  \vspace{-1.4\baselineskip}
    \caption{(a)-(b) Alternative passive mechanical equivalents of SDEA under VSIC during spring  rendering when controllers are~P, and (c)-(d) Realization of S(D)EA under VSIC during null impedance rendering when  controllers are~PI.}
      \vspace{-1\baselineskip}
    \label{fig:alternativerealall2}
  \end{figure}

Figures~\ref{fig:alternativerealall2}a--b depict alternative passive mechanical equivalents for the impedance in Eqn.~(\ref{eqn:SDEApsringPP}) for SDEA under VSIC during spring rendering when both controllers are~P. Here, Figures~\ref{fig:alternativerealall2}a and~\ref{fig:alternativerealall2}b complement each other to provide the same sufficient conditions as presented for Table~\ref{fig:allresults}h.  While Bott-Duffin theorem~\cite{bott1949impedance} establishes that ideal transformers (levers) can be avoided in non-minimal physical realizations, we present Figures~\ref{fig:alternativerealall2}a and~\ref{fig:alternativerealall2}b as a set of alternative minimal realizations, since we prioritize minimality of the realizations, the feasibility conditions of these two realizations complement each other to recover the necessary and sufficient conditions for the passivity of Eqn.~(\ref{eqn:SDEApsringPP}), and the use of a lever to change direction provides an understanding on how negative values of fundamental elements (e.g., $k_{1n}$ and $c_{4n}$) can be avoided. % in feasible realizations.

%Figures~\ref{fig:alternativerealall2}a--b depict alternative passive physical realizations of SEA under VSIC during null impedance rendering when the force controller is P and the motion controller is~PI. 
%In particular,  Table~\ref{fig:allresults}b and Figures~\ref{fig:alternativerealall2}a and~\ref{fig:alternativerealall2}b present the realizations for the impedance transfer function in Eqn.~(\ref{eqn:SEAnullPIP}).
%\sout{Section~\ref{sec:SEApassivity} presents feasibility conditions for Table~\ref{fig:allresults}b and prove that they establish a set of sufficient conditions for passivity. Figure~\ref{fig:alternativerealall2}a and~\ref{fig:alternativerealall2}b are alternative realizations, where the serial inerter-damper term introduced due to the integral controller is realized through a serial spring-damper. The feasibility conditions for these realizations provide a different set of sufficient conditions for passivity. While Bott-Duffin theorem~\cite{bott1949impedance} establishes that ideal transformers (levers) can be avoided in non-minimal physical realizations, we present Figures~\ref{fig:alternativerealall2}a and~\ref{fig:alternativerealall2}b as a set of alternative minimal realizations, since the feasibility conditions of these two realizations complement each other to recover the necessary and sufficient conditions for the passivity of Eqn.~(\ref{eqn:SEAnullPIP})  and the use of a lever to change direction provides an understanding on how negative values of fundamental elements (e.g., $k_{1n}$ and $c_{4n}$) can be avoided in feasible realizations.}

Realizations become more complicated as controllers become more involved, making their interpretation harder. For instance,  Figures~\ref{fig:alternativerealall2}c--d present passive physical realizations for SEA and SDEA under VSIC during null impedance rendering, when both controllers are PI. As the realizations become more complicated, the feasibility conditions for the realizations are likely to cover a smaller range of passive system parameters; hence, conclusions drawn from such realizations become valid for a limited range. Accordingly, it is preferable to utilize the simplest models competent to represent the essential dynamic behavior, as recommended in~\cite{Colgate_realizations}.

\vspace{-.7\baselineskip}

\section*{Acknowledgment} %\vspace{-1mm}
%We thank anonymous reviewers for their constructive feedback.  
This work has been partially supported under T\"{U}B\.{I}TAK Grants~216M200 and~23AG003.
%% Use plainnat to work nicely with natbib. 
\vspace{-3mm}

\appendices
\section*{Appendix A}
\label{appendix:A}

\begin{proof} First, note that asymptotic stability of the inner loop imposes {\small $(B_m\! +\! G_m) > 0$}. Next, according to Theorem~\ref{theorem:positivereallness}; 
\begin{itemize}
\item[(1)] \textit{\small $Z(s)$ has no poles in the right half plane:}
Invoking Lemma~\ref{lemma:stability} imposes {\small ${\left( \alpha\!+1 \right)}\,{\left(B_m \!+G_m +B_f(\alpha\!+1) \right)}\!\! \ge \!\! 0$}. Accordingly, $Z_{Voigt}^{SDEA^{P{\text -}P}}(s)$ has no roots in the open right half plane, if {\small $(\alpha+1)$ and $B_m +G_m +B_f(\alpha+1)$} are non-negative.

\item[(2)] \textit{\small Any poles of Z(s) on the imaginary axis are simple with positive and real residues:} If  {\small $(\alpha+1)$} is positive, then only possible root on the imaginary axis is $s=0$ as long as physical parameters are positive and inner motion loop is asymptotically stable. For the pole at $s=0$, the residue equals to {\small $\frac{\alpha}{(\alpha+1)} \, K_{ref}$} which should be positive. If {\small $(\alpha+1)=0$}, then the output impedance transfer function has double roots and Condition~3 of Theorem~\ref{theorem:positivereallness} is violated due to double poles at $s=0$.  Hence, when {\small $(\alpha+1)=0$}, passive Voigt models cannot be rendered.

\item[(3)] \textit{\small $Re[Z(jw)]  \ge \! 0$ for all $w$:} The sign of {\small $Re[Z_{voigt}^{SDEA^{P{\text -}P}}\!\!(jw)]$} can be checked by the sign of the test polynomial {\small $H(jw)= d_6 w^6 + d_4 w^4 + d_2 w^2$} from Lemma~\ref{lemma:positivereal}, where  \vspace{-8mm}

 \footnotesize
\begin{align}  \vspace{-6mm}
   \!\!\!\!\!\! d_2 =&  K^2 \,{\left(\alpha \!+\!1\right)}\,{\left(B_m \!+\! G_m\! +\!B_{ref} \,\alpha \right)}\!-{(B_m\! + \!G_m)} \, K\, K_{ref} \,\alpha   \\ 
   \!\!\!\!\!\! d_4 = & B_f \,{\left(B_m +G_m +B_{ref} \, \alpha \right)}\,{\left[B_m +G_m +B_f \, (\alpha+1) \right]} \nonumber \\ 
   \!\!\!\!\!\!  & - (B_{ref} \,K + B_f \, K_{ref}) \, J_m \, \alpha\\
   \!\!\!\!\!\!    d_6 = & B_f \,{J_m}^2
\end{align} \normalsize  \vspace{-6mm}

\noindent Applying Lemma~\ref{lemma:sturm}, and noting that $d_6$ is positive, since $B_f$ is positive, the following constraint is imposed by the non-negativeness~of~$d_2$: \vspace{-2mm}

 \footnotesize
\begin{equation}  
K\, \ge K_{ref} \, \frac{\alpha}{(\alpha+1)} \, \frac{B_m+G_m}{ B_m + G_m +   B_{ref} \, \alpha}
%K_{ref} \le  \left( \frac{{\left(\alpha +1\right)}}{\alpha }+\frac{B_{ref} \,{\left(\alpha +1\right)}}{B_m +G_m } \right)
    \label{eqn:SDEAvoigtPPd2}
\end{equation} \normalsize \vspace{-1.5mm}
\noindent The last necessary and sufficient condition reads as: \vspace{-3mm}

 \scriptsize
\begin{eqnarray}
\!\!\!\!\!\!\!\!\!\!\!\!\!\! && \!\!\!\!\!\!\!\!\!\!\!\!\!\!  -2 \, J_m\sqrt{B_f \, K \, \left[ (B_m \!+G_m \!+ B_{ref} \, \alpha ) \, K \,{(\alpha +1)} - (B_m\! + G_m) \, K_{ref} \, \alpha \right]} \nonumber \\
\!\!\!\!\!\!\!\!\!\!\!\!\!\! &&  \le  B_f \,{\left(B_m \!+ G_m \!+ B_{ref} \, \alpha \right)}\, \left[ B_m\! + G_m\! + B_f \, (\alpha+1) \right]   \\
\!\!\!\!\!\!\!\!\!\!\!\!\!\! &&   -(B_{ref} \,K \!+ B_f \, K_{ref}) \, J_m \, \alpha  \nonumber
\end{eqnarray} \normalsize
%
%\noindet where \\ 
%$\gamma \!= \! -B_f \, K \, {J_m }^2 \,({(G_m + B_m)} \, K_{ref} \, \alpha - K \,{(\alpha +1)}\,(B_m +G_m + B_{ref} \, \alpha ) )$ \normalsize
\end{itemize}
\vspace{-1.5\baselineskip}
\end{proof}
\vspace{-3mm}

%Table~\ref{table:massdamperadd} shows the additional serial-inerter damper values for various $G_t$, $G_m$, $I_m$, and $K_{ref}$ for SEA null impedance rendering when controllers are PI-P and SDEA spring rendering when controllers are P-P. We can see that $b_{1n}$ and $c_{2n}$ decrease or same with higher gains. $b_{1s}$ decreases with higher $G_t$, slightly increases with higher $K_{ref}$, and stays same with higher $G_m$. $c_{1s}$ decreases with higher $G_t$, increases with higher $G_m$ and $K_{ref}$. 
%\vspace{-2mm}
%\vspace{-.75\baselineskip}
%\bigskip
\bibliographystyle{IEEEtran} %\vspace{-1.5mm}

\begin{thebibliography}{10}
\providecommand{\url}[1]{#1}
\csname url@samestyle\endcsname
\providecommand{\newblock}{\relax}
\providecommand{\bibinfo}[2]{#2}
\providecommand{\BIBentrySTDinterwordspacing}{\spaceskip=0pt\relax}
\providecommand{\BIBentryALTinterwordstretchfactor}{4}
\providecommand{\BIBentryALTinterwordspacing}{\spaceskip=\fontdimen2\font plus
\BIBentryALTinterwordstretchfactor\fontdimen3\font minus \fontdimen4\font\relax}
\providecommand{\BIBforeignlanguage}[2]{{%
\expandafter\ifx\csname l@#1\endcsname\relax
\typeout{** WARNING: IEEEtran.bst: No hyphenation pattern has been}%
\typeout{** loaded for the language `#1'. Using the pattern for}%
\typeout{** the default language instead.}%
\else
\language=\csname l@#1\endcsname
\fi
#2}}
\providecommand{\BIBdecl}{\relax}
\BIBdecl

\bibitem{colgate_hogan_1988}
J.~E. Colgate and N.~Hogan, ``Robust control of dynamically interacting systems,'' \emph{Int. Journal of Control}, vol.~48, no.~1, pp. 65–--88, 1988.

\bibitem{howard1990joint}
R.~D. Howard, ``Joint and actuator design for enhanced stability in robotic force control,'' Ph.D. dissertation, MIT, 1990.

\bibitem{pratt1995series}
G.~A. Pratt and M.~M. Williamson, ``Series elastic actuators,'' in \emph{IEEE/RSJ Int. Conf. on Intel. Rob. and Sys.}, vol.~1, 1995, pp. 399--406.

\bibitem{robinson1999series}
D.~W. Robinson, J.~E. Pratt, D.~J. Paluska, and G.~A. Pratt, ``Series elastic actuator development for a biomimetic walking robot,'' in \emph{IEEE/ASME Int. Conf. on Adv. Intel. Mech.}, 1999, pp. 561--568.

\bibitem{eppinger_seering}
S.~Eppinger and W.~Seering, ``Understanding bandwidth limitations in robot force control,'' in \emph{IEEE Int. Conf. on Rob. and Auto.}, vol.~4, 1987, pp. 904--909.

\bibitem{newman_1992}
W.~S. Newman, ``Stability and performance limits of interaction controllers,'' \emph{J. of Dynamic Sys., Meas., and Cont.}, vol. 114, no.~4, pp. 563–--570, 1992.

\bibitem{kenanoglu2023}
C.~U. Kenanoglu and V.~Patoglu, ``A fundamental limitation of passive spring rendering with series elastic actuation,'' \emph{IEEE Trans. on Haptics}, vol.~16, no.~4, pp. 456--462, 2023.

\bibitem{oblak2011design}
J.~Oblak and Z.~Matja{\v{c}}i{\'c}, ``Design of a series visco-elastic actuator for multi-purpose rehabilitation haptic device,'' \emph{J. of Neuroeng. and Rehab.}, vol.~8, no.~1, pp. 1--14, 2011.

\bibitem{hurst2004series}
J.~Hurst, A.~Rizzi, and D.~Hobbelen, ``Series elastic actuation: Potential and pitfalls,'' in \emph{Int. Conf. on Climbing and Walking Rob.}, 2004.

\bibitem{garcia2011combining}
E.~Garcia, J.~C. Arevalo, G.~Mu{\~n}oz, and P.~Gonzalez-de Santos, ``Combining series elastic actuation and magneto-rheological damping for the control of agile locomotion,'' \emph{Rob. and Auto. Sys.}, vol.~59, no.~10, pp. 827--839, 2011.

\bibitem{laffranchi2011compact}
M.~Laffranchi, N.~Tsagarakis, and D.~G. Caldwell, ``A compact compliant actuator (compact) with variable physical damping,'' in \emph{IEEE Inter. Conf. on Rob. and Auto.}, 2011, pp. 4644--4650.

\bibitem{kim2017enhancing}
M.~J. Kim, A.~Werner, F.~C. Loeffl, and C.~Ott, ``Enhancing joint torque control of series elastic actuators with physical damping,'' in \emph{IEEE Int. Conf. on Rob. and Auto.}, 2017, pp. 1227--1234.

\bibitem{pratt2004}
G.~Pratt, P.~Willisson, C.~Bolton, and A.~Hofman, ``Late motor processing in low-impedance robots: impedance control of series-elastic actuators,'' in \emph{American Control Conference}, 2004, pp. 3245--3251.

\bibitem{Wyeth2006}
G.~Wyeth, ``Control issues for velocity sourced series elastic actuators,'' in \emph{Australasian Conf. on Rob. and Auto.}, 2006, pp. 1--6.

\bibitem{Wyeth2008}
------, ``Demonstrating the safety and performance of a velocity sourced series elastic actuator,'' in \emph{IEEE Inter. Conf. on Rob. and Auto.}, 2008, pp. 3642--3647.

\bibitem{vallery2007passive}
H.~Vallery, R.~Ekkelenkamp, H.~Van Der~Kooij, and M.~Buss, ``Passive and accurate torque control of series elastic actuators,'' in \emph{IEEE/RSJ Int. Conf. on Intel. Rob. and Sys.}, 2007, pp. 3534--3538.

\bibitem{vallery2008compliant}
H.~Vallery, J.~Veneman, E.~Van~Asseldonk, R.~Ekkelenkamp, M.~Buss, and H.~Van Der~Kooij, ``Compliant actuation of rehabilitation robots,'' \emph{IEEE Rob. and Auto. Mag.}, vol.~15, no.~3, pp. 60--69, 2008.

\bibitem{tagliamonte2014passivity}
N.~L. Tagliamonte and D.~Accoto, ``Passivity constraints for the impedance control of series elastic actuators,'' \emph{Inst. of Mech. Eng., Part I: J. of Sys. and Cont. Eng.}, vol. 228, no.~3, pp. 138--153, 2014.

\bibitem{calanca2017impedance}
A.~Calanca, R.~Muradore, and P.~Fiorini, ``Impedance control of series elastic actuators: Passivity and acceleration-based control,'' \emph{Mechatronics}, vol.~47, pp. 37--48, 2017.

\bibitem{Tosun2020}
F.~E. Tosun and V.~Patoglu, ``Necessary and sufficient conditions for the passivity of impedance rendering with velocity-sourced series elastic actuation,'' \emph{IEEE Trans. on Robotics}, vol.~36, no.~3, pp. 757--772, 2020.

\bibitem{Kenanoglu2023b}
O.~T. Kenanoglu, C.~U. Kenanoglu, and V.~Patoglu, ``Effect of low-pass filtering on passivity and rendering performance of series elastic actuation,'' \emph{IEEE Trans. on Haptics}, vol.~16, no.~4, pp. 567--573, 2023.

\bibitem{Otaran2021}
A.~Otaran, O.~Tokatli, and V.~Patoglu, ``Physical human-robot interaction using {HandsOn-SEA: An} educational robotic platform with series elastic actuation,'' \emph{IEEE Trans. on Haptics}, vol.~14, no.~4, pp. 922--929, 2021.

\bibitem{Calanca2018_ff}
A.~Calanca and P.~Fiorini, ``A rationale for acceleration feedback in force control of series elastic actuators,'' \emph{IEEE Trans. on Robotics}, vol.~34, no.~1, pp. 48--61, 2018.

\bibitem{Kenanoglu2022}
C.~U. Kenanoglu and V.~Patoglu, ``Passivity of series elastic actuation under model reference force control during null impedance rendering,'' \emph{IEEE Trans. on Haptics}, vol.~15, no.~1, pp. 51--56, 2022.

\bibitem{focchi2016robot}
M.~Focchi, G.~A. Medrano-Cerda, T.~Boaventura, M.~Frigerio, C.~Semini, J.~Buchli, and D.~G. Caldwell, ``Robot impedance control and passivity analysis with inner torque and velocity feedback loops,'' \emph{Control Theory and Tech.}, vol.~14, no.~2, pp. 97--112, 2016.

\bibitem{Mengilli2020}
\BIBentryALTinterwordspacing
U.~Mengilli, U.~Caliskan, Z.~O. Orhan, and V.~Patoglu, ``Two-port analysis of stability and transparency in series damped elastic actuation,'' \emph{CoRR}, 2020. [Online]. Available: \url{https://arxiv.org/abs/2011.00664}
\BIBentrySTDinterwordspacing

\bibitem{mengilli2021passivity}
U.~Mengilli, Z.~O. Orhan, U.~Caliskan, and V.~Patoglu, ``Passivity of series damped elastic actuation under velocity-sourced impedance control,'' in \emph{IEEE World Haptics Conf.}, 2021, pp. 379--384.

\bibitem{Colgate_realizations}
E.~Colgate and N.~Hogan, ``An analysis of contact instability in terms of passive physical equivalents,'' in \emph{Int. Conf. on Rob. and Auto.}, vol.~1, 1989, pp. 404--409.

\bibitem{smith2002synthesis}
M.~C. Smith, ``Synthesis of mechanical networks: the inerter,'' \emph{IEEE Trans. on Automatic Control}, vol.~47, no.~10, pp. 1648--1662, 2002.

\bibitem{Chen_missing}
M.~Z. Chen, C.~Papageorgiou, F.~Scheibe, F.-c. Wang, and M.~C. Smith, ``The missing mechanical circuit element,'' \emph{IEEE Circ. and Sys. Mag.}, vol.~9, no.~1, pp. 10--26, 2009.

\bibitem{foster1924reactance}
R.~M. Foster, ``A reactance theorem,'' \emph{Bell System Technical Journal}, vol.~3, no.~2, pp. 259--267, 1924.

\bibitem{brune1931synthesis}
O.~Brune, ``Synthesis of a finite two-terminal network whose driving-point impedance is a prescribed function of frequency,'' Ph.D. dissertation, MIT, 1931.

\bibitem{bott1949impedance}
R.~Bott and R.~Duffin, ``Impedance synthesis without use of transformers,'' \emph{Journal of Applied Physics}, vol.~20, no.~8, pp. 816--816, 1949.

\bibitem{chen2009restricted}
M.~Z. Chen and M.~C. Smith, ``Restricted complexity network realizations for passive mechanical control,'' \emph{IEEE Trans. on Auto. Control}, vol.~54, no.~10, pp. 2290--2301, 2009.

\bibitem{chen2012realization}
M.~Z. Chen, K.~Wang, Y.~Zou, and J.~Lam, ``Realization of a special class of admittances with one damper and one inerter,'' in \emph{IEEE Conf. on Decision and Control}, 2012, pp. 3845--3850.

\bibitem{chen2013realization}
------, ``Realization of a special class of admittances with one damper and one inerter for mechanical control,'' \emph{IEEE Trans. on Auto. Control}, vol.~58, no.~7, pp. 1841--1846, 2013.

\bibitem{chen2013realizations}
M.~Z. Chen, K.~Wang, Z.~Shu, and C.~Li, ``Realizations of a special class of admittances with strictly lower complexity than canonical forms,'' \emph{IEEE Trans. on Circ. and Sys. I}, vol.~60, no.~9, pp. 2465--2473, 2013.

\bibitem{chen2015realization}
M.~Z. Chen, K.~Wang, Y.~Zou, and G.~Chen, ``Realization of three-port spring networks with inerter for effective mechanical control,'' \emph{IEEE Trans. on Auto. Control}, vol.~60, no.~10, pp. 2722--2727, 2015.

\bibitem{Kalman2010}
R.~Kalman, \emph{Perspectives in Mathematical System Theory, Control, and Signal Processing}.\hskip 1em plus 0.5em minus 0.4em\relax Berlin, Heidelberg: Springer, 2010, vol. 398, ch. Old and new directions of research in system theory, p.~3.

\bibitem{jiang2012realization}
J.~Z. Jiang and M.~C. Smith, ``Series-parallel six-element synthesis of biquadratic impedances,'' \emph{IEEE Trans. on Circ. and Sys. I}, vol.~59, no.~11, pp. 2543--2554, 2012.

\bibitem{Hughes_2014_realization}
T.~H. Hughes and M.~C. Smith, ``On the minimality and uniqueness of the bott–duffin realization procedure,'' \emph{IEEE Trans. on Auto. Control}, vol.~59, no.~7, pp. 1858--1873, 2014.

\bibitem{Hughes_2020}
T.~H. Hughes, ``Minimal series–parallel network realizations of bicubic impedances,'' \emph{IEEE Trans. on Auto. Control}, vol.~65, no.~12, pp. 4997--5011, 2020.

\bibitem{Morelli_book_2019}
A.~Morelli and M.~C. Smith, \emph{Passive Network Synthesis: An Approach to Classification}.\hskip 1em plus 0.5em minus 0.4em\relax SIAM, 2019.

\bibitem{Hughes_survey_2018}
S.~M. Hughes~T.H., Morelli~A., \emph{Electrical Network Synthesis: A Survey of Recent Work}, ser. Lecture Notes in Control and Inf. Sci.\hskip 1em plus 0.5em minus 0.4em\relax Springer, 2018, ch. Emerg. App. of Contr. and Sys. Theory, pp. 281--293.

\bibitem{Hannaford1989}
B.~Hannaford, ``A design framework for teleoperators with kinesthetic feedback,'' \emph{IEEE Trans. on Rob. and Auto.}, vol.~5, no.~4, pp. 426--434, 1989.

\bibitem{Hashtrudi-Zaad2001}
K.~Hashtrudi-Zaad and S.~E. Salcudean, ``Analysis of control architectures for teleoperation systems with impedance/admittance master-slave manipulators,'' \emph{Int. J. of Rob. Res.}, vol.~20, no.~6, pp. 419--445, 2001.

\bibitem{Colgatezwidth94}
J.~Colgate and J.~Brown, ``Factors affecting the {Z-width} of a haptic display,'' in \emph{IEEE Int. Conf. on Rob. and Auto.}, 1994, pp. 3205--3210.

\bibitem{mehling2005increasing}
J.~S. Mehling, J.~E. Colgate, and M.~A. Peshkin, ``Increasing the impedance range of a haptic display by adding electrical damping,'' in \emph{IEEE World Haptics Conf.}, 2005, pp. 257--262.

\bibitem{colonnese2015rendered}
N.~Colonnese, A.~F. Siu, C.~M. Abbott, and A.~M. Okamura, ``Rendered and characterized closed-loop accuracy of impedance-type haptic displays,'' \emph{IEEE Trans. on Haptics}, vol.~8, no.~4, pp. 434--446, 2015.

\bibitem{tokatli2015stability}
O.~Tokatli and V.~Patoglu, ``Stability of haptic systems with fractional order controllers,'' in \emph{IEEE/RSJ Int. Conf. on Intel. Rob. and Sys.}, 2015, pp. 1172--1177.

\bibitem{tokatli_patoglu_2018}
------, ``Using fractional order elements for haptic rendering,'' \emph{Springer Adv. Rob. Res.}, pp. 373--–388, 2018.

\bibitem{aydin2018stable}
Y.~Aydin, O.~Tokatli, V.~Patoglu, and C.~Basdogan, ``Stable physical human-robot interaction using fractional order admittance control,'' \emph{IEEE Trans. on Haptics}, vol.~11, no.~3, pp. 464--475, 2018.

\bibitem{Yusuf_multi_2020}
------, ``A computational multicriteria optimization approach to controller design for physical human-robot interaction,'' \emph{IEEE Trans. on Robotics}, vol.~36, no.~6, pp. 1791--1804, 2020.

\bibitem{Haykin70}
S.~Haykin, \emph{Active network theory}.\hskip 1em plus 0.5em minus 0.4em\relax Addison-Wesley Pub. Co., 1970.

\bibitem{supplementary}
\BIBentryALTinterwordspacing
C.~U. Kenanoglu and V.~Patoglu, ``Supplementary document for passive realizations of series elastic actuation,'' Sabanci University, Tech. Rep., 2022. [Online]. Available: \url{https://tinyurl.com/u5f2e5ev}
\BIBentrySTDinterwordspacing

\bibitem{Kamadan2017}
A.~Kamadan, G.~Kiziltas, and V.~Patoglu, ``Co-design strategies for optimal variable stiffness actuation,'' \emph{IEEE/ASME Trans. on Mech.}, vol.~22, no.~6, pp. 2768--2779, 2017.

\bibitem{Kamadan2019}
------, ``A systematic design selection methodology for system-optimal compliant actuation,'' \emph{Robotica}, vol.~37, no.~4, pp. 656--–674, 2019.

\bibitem{Weir2008}
D.~W. Weir, J.~E. Colgate, and M.~A. Peshkin, ``Measuring and increasing z-width with active electrical damping,'' \emph{Symp. on Haptic Interfaces for Virtual Environment and Teleoperator Systems}, pp. 169--175, 2008.

\bibitem{caliskan2018series}
U.~Caliskan, A.~Apaydin, A.~Otaran, and V.~Patoglu, ``A series elastic brake pedal to preserve conventional pedal feel under regenerative braking,'' in \emph{IEEE/RSJ Int. Conf. on Intel. Rob. and Sys.}, 2018.

\bibitem{caliskan2020efficacy}
U.~Caliskan and V.~Patoglu, ``Efficacy of haptic pedal feel compensation on driving with regenerative braking,'' \emph{IEEE Trans. on Haptics}, vol.~13, no.~1, pp. 175--182, 2020.

\bibitem{colgate_1988}
J.~E. Colgate, ``The control of dynamically interacting systems,'' Ph.D. dissertation, MIT, 1988.

\bibitem{Losey2017}
D.~P. {Losey} and M.~K. {O'Malley}, ``Effects of discretization on the k-width of series elastic actuators,'' in \emph{IEEE Int. Conf. on Rob. and Auto.}, 2017, pp. 421--426.

\end{thebibliography}
\end{document}